%% file: Thesis.tex
\documentclass[bs]{osudissert96}

\usepackage{natbib}

\usepackage{pbox}
\usepackage{xspace}
\newcommand{\cough}{\texttt{COUGH}\xspace}

\usepackage{microtype}
\usepackage{times}
\usepackage{latexsym}
\usepackage{microtype}
\usepackage{graphicx}
\usepackage{multirow}
\usepackage{soul}
\usepackage{amsmath}
\usepackage{amssymb}
\usepackage{enumitem}
\usepackage{booktabs}
\usepackage[normalem]{ulem}
\useunder{\uline}{\ul}{}
\usepackage{setspace}
\usepackage[belowskip=1pt,aboveskip=4pt]{caption}
\renewcommand\vec{\mathbf}
\usepackage{tablefootnote}
\usepackage{xcolor}
\usepackage{algorithm}
\usepackage{algpseudocode}
\usepackage{diagbox}

\newcommand{\nop}[1]{}

\def\BibTeX{{\rm B\kern-.05em{\sc i\kern-.025em b}\kern-.08em
    T\kern-.1667em\lower.7ex\hbox{E}\kern-.125emX}}

\renewcommand\typesetChapterTitle[1]{#1}

\begin{document}

%
%

\author{Xinliang (Frederick) Zhang}
\title{Towards More Robust Natural Language Understanding}
\unit{Computer Science and Engineering}

\advisorname{Dr. Huan Sun}
\member{Dr. Marie-Catherine de Marneffe}

%
%

\maketitle

%
%
%
%
%

\disscopyright

%
%

\begin{abstract}
  \input{abstract}

\end{abstract}

%
%
%


%
%

\dedication{\textit{Dedicated to my parents.}}

%
%

\include{ack}

\include{vita}

%
%

\tableofcontents
\listoftables
\listoffigures

%
%

\include{ch1.Intro}

\include{ch2.CQA}   
\include{ch4.NLI}
\include{ch3.FAQ}

\include{ch5.Conclusion}

%
%

\appendix
\include{app1}

%
%


\end{document}

%% file: abstract.tex




Natural Language Understanding (NLU) is a branch of Natural Language Processing (NLP) that uses intelligent computer software to understand texts that encode human knowledge. Recent years have witnessed notable progress across various NLU tasks with deep learning techniques, especially with pretrained language models. Besides proposing more advanced model architectures, constructing more reliable and trustworthy datasets also plays a huge role in improving NLU systems, without which it would be impossible to train a decent NLU model. It's worth noting that the human ability of understanding natural language is flexible and robust. On the contrary, most of existing NLU systems fail to achieve desirable performance on out-of-domain data or struggle on handling challenging items (e.g., inherently ambiguous items, adversarial items) in the real world. Therefore, in order to have NLU models understand human language more effectively, it is expected to prioritize the study on robust natural language understanding.

In this thesis, we deem that NLU systems are consisting of two components: NLU models and NLU datasets. As such, we argue that, to achieve robust NLU, the model architecture/training and the dataset are equally important. Specifically, we will focus on three NLU tasks to illustrate the robustness problem in different NLU tasks and our contributions (i.e., novel models and new datasets) to help achieve more robust natural language understanding. The major technical contributions of this thesis are: 

\begin{itemize}[leftmargin=2.8em,noitemsep,topsep=0pt,parsep=0pt,partopsep=0pt]
    \item[1. ] We study how to utilize diversity boosters (e.g., beam search \& QPP) to help neural question generator synthesize diverse QA pairs, upon which a Question Answering (QA) system is trained to improve the generalization on the unseen target domain. It's worth mentioning that our proposed QPP (question phrase prediction) module, which predicts a set of valid question phrases given an answer evidence, plays an important role in improving the cross-domain generalizability for QA systems. Besides, a target-domain test set is constructed and approved by the community to help evaluate the model robustness under the cross-domain generalization setting. 
    \item[2. ] We investigate inherently ambiguous items in Natural Language Inference, for which annotators don’t agree on the label. Ambiguous items are overlooked in the literature but often occurring in the real world. We build an ensemble model, AAs (Artificial Annotators), that simulates underlying annotation distribution to effectively identify such inherently ambiguous items. Our AAs are better at handling inherently ambiguous items since the model design captures the essence of the problem better than vanilla model architectures.
    \item[3. ] We follow a standard practice to build a robust dataset for FAQ retrieval task, \cough. In our dataset analysis, we show how \cough better reflects the challenge of FAQ retrieval in the real situation than its counterparts. The imposed challenge will push forward the boundary of research on FAQ retrieval in real scenarios.
\end{itemize}

Moving forward, the ultimate goal for robust natural language understanding is to build NLU models which can behave humanly. That is, it’s expected that robust NLU systems are capable to transfer the knowledge from training corpus to unseen documents more reliably and survive when encountering challenging items even if the system doesn’t know a priori of users’ inputs.

%% file: ack.tex
\begin{acknowledgements}

I feel incredibly fortunate to have Dr. Huan Sun as my advisor, without whom nothing in this thesis is possible. I would like to express my sincere gratitude to her for critiquing my work and my ideas in a constructive way. Her vision and rigorous research attitudes have shaped my thoughts. I am always indebted to her for guiding me all the way, for her contagious energy, for being so supportive and caring about students.

I owe a great debt of gratitude to Dr. Marie-Catherine de Marneffe for her countless help. I am super grateful for her many invaluable insights and suggestions on my work. She has been so generous with her time, reading, reviewing and commenting on many of my writings. I am always thankful for her positive encouragement and praise.

It's also my great privilege to collaborate with my friends, lab-mates at SunLab and my past mentor: Xiang Yue, Ziyu Yao, Heming Sun, Emmett Jesrani and Dr. Chen Chen. 

I appreciate the help from anyone who helped me along my education journey: Hangzhou Jindu Tianchang Elementary School, Hangzhou Caihe Experimental School, Hangzhou Xuejun High School, Sichuan University and Ohio State University. I am especially thankful to my Chinese friends, who always compliment me and cheer me up no matter what.

Last but not the least, my deepest gratitude, without any doubt, goes to my parents, Rongchang and Hangjuan. They gave birth to me, raised me up, set good examples for me, and taught me tremendously many invaluable lessons. I wouldn't become who I am without their trust and support. All in all, thanks for their unconditional love and upbringing.

\end{acknowledgements}

%% file: vita.tex
\begin{vita}
\dateitem{2021-}{Ph.D. in Computer Science and Engineering, University of Michigan.}

\dateitem{2018-2021}{B.S. in Computer Science and Engineering \& Industrial and Systems Engineering, The Ohio State University.}

\dateitem{2016-2018}{B.E. in Industrial Engineering, Sichuan University.}

\begin{publist}


\researchpubs

\pubitem{\textbf{Xinliang~Frederick Zhang} and Marie{-}Catherine de~Marneffe.
\newblock Identifying inherent disagreement in natural language inference.
\newblock In {\em  NAACL 2021.} 2021.}

\pubitem{\textbf{Xinliang~Frederick Zhang}, Heming Sun, Xiang Yue, Simon Lin, and Huan Sun.
\newblock {COUGH}: A challenge dataset and models for {COVID-19 FAQ} retrieval.
\newblock In {\em  EMNLP 2021.} 2021.}

\pubitem{Xiang Yue*, \textbf{Xinliang~Frederick Zhang*}, Ziyu Yao, Simon Lin, and Huan Sun.
\newblock Clini{QG4QA}: Generating diverse questions for domain adaptation of clinical question answering.
\newblock In {\em IEEE BIBM 2021}. 2021. (*equal contributions)}

%

\end{publist}

\begin{fieldsstudy}

\majorfield*



\end{fieldsstudy}

\end{vita}

%% file: ch1.Intro.tex
\chapter{Introduction}
\label{intro}

\section{Natural Language Understanding (NLU)}
Have you ever asked: ``Siri, how is the weather today?'', ``Cortana, what is the best spot for hiking in Columbus?'' or "Xiaoice, could you tell me how's traffic outside?''.  If so, you have experienced receiving a data-supported answer from your personalized AI assistant. A natural question that people would ask is how can the agent understand an utterance and intents and generate a relevant response. The answer is Natural Language Understanding.

Natural Language Understanding (NLU) is a branch of Natural Language Processing (NLP) in the area of Artificial Intelligence (AI) that uses intelligent computer software to understand texts that encode human knowledge. Some representative NLU applications (and there are way more) are: Automated Reasoning, Question Answering, Text Categorization, Large-scale Content Analysis, Information Retrieval and Textual Entailment. NLU is generally considered an AI-hard problem (i.e., a problem that is hard to be solved by AI systems) \citep{Yampolskiy_AI}. NLU is an AI-hard problem mainly because the nature of human language (e.g., ambiguity) makes NLU difficult. For example, given the following sentence ``when the hammer hit the glass table, it shattered'',\footnote{https://www.colorado.edu/earthlab/2020/02/07/what-natural-language-processing-and-why-it-hard.} humans know that it is the glass table that shattered but not the hammer. This is because our prior knowledge let us know what glass is and that glass can shatter easily. However, coreference resolution is still a challenging task for NLU models, and thus, NLU systems still have difficulties figuring out which one of these two objects shatters.

Recent years have witnessed notable progress across various Natural Language Understanding tasks, especially after entering the deep learning era in 2012. Deep learning approaches quickly outperformed statistical learning methods by a large margin on many NLU tasks. As today, neural network-based NLP models have reached many new milestones (e.g., model performance comes close to or surpasses the level of non-expert humans) and have become the dominating approach for NLP tasks. Typical neural network-based NLP models/algorithms are RNN \citep{RNN}, LSTM \citep{LSTM}, GRU \citep{GRU}, Seq2Seq \citep{SutskeverVL14}, attention mechanism \citep{luong2015effective} and Transformer \citep{1706.03762}. Recently, pretrained language models, such as GPT \citep{radford2018improving} and BERT \citep{devlin2019bert}, have dramatically altered the NLP landscape and marked new records on the majority of NLU tasks. However, the neural NLP models work well for supervised tasks in which there is abundant labeled data for learning, but still perform poorly for low-resource and cross-domain tasks where the training data is insufficient and the test data is from different domains, respectively.

Besides more advanced model architectures, reliable and trustworthy datasets also play a huge role in improving NLU systems. Without a decent dataset, it would be challenging to train a machine learning model, not to mention carrying out a valid evaluation. As such, comprehensive evaluation benchmarks, aggregating datasets of multiple NLU tasks, emerged in the past few years such as GLUE \citep{WangSMHLB19} and SuperGLUE \citep{WangPNSMHLB19}. They are diagnostic datasets designed to evaluate and analyze model performance with respect to a wide range of linguistic phenomena found in human language. 

The human ability of understanding natural language is flexible and robust. Therefore, human capability of understanding multiple language tasks simultaneously and transferring the knowledge to unseen documents is mostly reliable. On the contrary, most of existing NLU models built on word/character levels are exclusively trained on a restricted dataset. These restricted datasets normally only characterize one particular domain or only include simple examples which might not well reflect the task difficulties in reality. Consequently, such models usually fail to achieve desirable performance on out-of-domain data or struggle on handling challenging items (e.g., inherently ambiguous items, adversarial items) in the real world. Moreover, machine learning algorithms are usually data-hungry and can easily malfunction when there is insufficient amount of training data.  Therefore, in order to have NLU models understand human language more effectively, it is expected to prioritize the study on robust natural language understanding.

\section{Robustness Problem in NLU}

In this thesis, we deem that NLU systems are consisting of two components: NLU models and NLU datasets. As such, we argue that, to achieve robust NLU, the model architecture/training and the dataset are equally important. If either component is weak, it would be hard to achieve full robustness. Therefore, in order to achieve full robustness in NLU, researchers are expected to implement robust models which then are trained on constructed robust datasets. In this thesis, we define robust models and robust datasets as follow:

\begin{itemize}[leftmargin=2.8em,noitemsep,topsep=0pt,parsep=0pt,partopsep=0pt]
    \item[1. ] \textbf{Robust models} are expected to be resistant to domain changes and resilient to challenging items (e.g., inherently ambiguous items, adversarial items).
    \item[2. ] \textbf{Robust datasets} are expected to reflect real-world challenges and encode knowledge that is difficult to be unraveled simply by surface-level\footnote{For example, the presence of ``not'' or ``bad'' doesn't always indicate a negative sentiment.} understanding.
\end{itemize}

In short, a truly robust NLU system is expected to be a robust model trained on robust datasets.

\noindent\textbf{Three NLU tasks for NLU robustness problem}

In the context of NLP, robustness is an umbrella term which could be interpreted differently from different angles. In this thesis, we will focus on three NLU tasks to illustrate the robustness problem in different NLU tasks and our contributions (i.e., novel models and new datasets) to help achieve more robust natural language understanding.

The first robustness problem that will be studied in this thesis is the cross-domain generalization.In Question Answering, most past work on open-domain were only testing models on in-domain data (\textit{source} domain), despite outperforming human performance. However, these well-performing models have a relatively weak generalizability, which is the crux of this robustness problem. That is, when such models are deployed on out-of-domain data (\textit{target} domain), their performances go down drastically, which is way behind human performance. Similar trend is also observed under the clinical setting where a model trained on one corpus may not generalize well to new clinical texts collected from different medical institutions \citep{yue2020CliniRC,cliniqg4qa2020}. In Chapter~\ref{CQA}, we will study how to utilize diversity boosters to help Question Generator (QG) synthesize diverse\footnote{``Diverse'' here means questions with different syntactic structures or different topics.} QA pairs, upon which a Question Answering system is trained to improve the generalization to the unseen target domain. 
We also construct a target-domain test set to help evaluate models' generalizability.

The second robustness problem that will be studied in this thesis is how to better handle inherently ambiguous items, one type of challenging items in reality. In sentiment analysis and textual entailment tasks, it has been observed that there are inherently ambiguous/disagreement\footnote{In this thesis, ``ambiguous'' and ``disagreement'' will be used interchangeably.} items for which annotators have different annotations \citep{Kenyon-Dean18, PavlickK19,zhang2021NLI}. These items were usually treated as noise and removed in the dataset construction phase, which is problematic. In Chapter~\ref{NLI}, we will investigate inherently ambiguous items,  which are overlooked in the literature but often occurring in the real world, in the NLI (Natural Language Inference) task. To this end, we build an ensemble model, AAs (Artificial Annotators), which simulates underlying annotation distribution by capturing the modes in annotations to effectively identify such inherently ambiguous items.

The third robustness problem that will be studied in this thesis is how to construct a reliable and challenging dataset (i.e., robust dataset). In textual entailment and FAQ retrieval tasks, common datasets (e.g., SNLI \citep{bowman-etal-2015-large} and MultiNLI \citep{williams-etal-2018-broad} for textual entailment; FAQIR \citep{Karan2016} and StackFAQ \citep{Karan2018} for FAQ retrieval) used for training and testing might not well characterize the real difficulties of respective tasks. In the aforementioned datasets, sentence lengths and language complexities are generally low,  styles are limited  and the search space is small. In Chapter~\ref{FAQ}, we will follow a standard practice to build a robust dataset for the FAQ Retrieval task. In our dataset analysis, we will also show how this dataset better reflects the challenge of FAQ Retrieval in the real situation than its counterparts.

We will conclude with recommendations for future work about how to better approach robustness problem in NLU in Chapter~\ref{conclusion}.

%% file: ch2.CQA.tex
\chapter{Clinical Question Answering}
\label{CQA}

\section{Introduction}
Clinical question answering (QA), which aims to automatically answer natural language questions based on clinical texts in Electronic Medical Records (EMR), has been identified as an important task to assist clinical practitioners \citep{patrick2012ontology,raghavan2018annotating,pampari2018emrqa,fan2019annotating, rawat2020entity}. Neural QA models in recent years \citep{chen2017reading, devlin2019bert} show promising results in this research. However, answering clinical questions still remains challenging in real-world scenarios because well-trained QA systems may not generalize well to new clinical contexts from a different institute or patient group. For example, \citet{yue2020CliniRC} pointed out when a clinical QA model trained on the emrQA \citep{pampari2018emrqa} dataset is deployed to answer questions on MIMIC-III clinical texts \citep{mimiciii}, its performance drops by around 30\% even on questions that are similar to those in training.

Most of the existing clinical QA datasets and setups focus on in-domain testing while leaving the generalization challenge under-explored. In this chapter, we propose to evaluate \textit{the performance of clinical QA models on target contexts and questions which may have different distributions from the training data}. Due to the lack of publicly-available clinical QA pairs for our proposed evaluation setting, we ask clinical experts to annotate a new test set on the sampled MIMIC-III \citep{mimiciii} clinical texts.

Inspired by recent work on question generation (QG) for improving QA performance in the open domain \citep{golub2017two,wang2019adversarial,ShakeriSZNNWNX20}, we implement an answer evidence extractor and a seq2seq-based QG model to synthesize QA pairs on target contexts to train a QA model. However, we do not observe that such QA models achieve better performance on our curated MIMIC-III QA set, compared with that trained on emrQA.  Our error analysis reveals that the automatic generation technique often falls short of generating questions that are \emph{diverse} enough to serve as useful training data for clinical QA models.

To this end, we investigate two kinds of approaches to diversify the generation. Inspired by \citet{IppolitoKSKC19} whio study various decoding-based methods, we pick the standard beam search as a representative of the decoding-based approach since it achieves satisfying performance in various generation tasks. On the other hand, another practice (topic-guided approach) is to have a diversification step followed by a conditional generation. In general, such techniques first decide question topics and then generate questions conditioned on selected topics \citep{kang2019let,cho2019mixture,liu2020asking}. Following the second approach, we propose a simple but effective question phrase prediction (QPP) module to diversify the generation. Specifically, QPP takes the extracted answer evidence as input and sequentially predicts potential question phrases (e.g., ``What treatment'', ``How often'') that signify what types of questions humans may ask about the answer evidence. Then, by directly forcing a QG model to produce specified question phrases at the beginning of the question generation process (both in training and inference),  QPP enables diverse questions to be generated.

Through comprehensive experiments, we demonstrate that when using QA pairs automatically synthesized by diverse QG, especially by the QPP-enhanced QG, we are able to boost QA performance by 4.5\%-9\% in terms of Exact Match (EM), compared with their counterparts directly trained on the source QA dataset (i.e., emrQA).

\section{Out-of-Domain Test Set}
Unlike open domain, there are very few publicly available QA datasets in the clinical domain. EmrQA dataset \citep{pampari2018emrqa}, which was generated based on medical expert-made question templates and existing annotations on n2c2 challenge datasets \citep{n2c2}, is a commonly adopted dataset for clinical reading comprehension. 

However, all the QA pairs in emrQA are based on n2c2 clinical texts and thus not suitable for our generalization setting. \citet{yue2020CliniRC} studied a similar problem and annotated a test set on MIMIC-III clinical texts \citep{mimiciii}. However, their test set is too small (only 50 QA pairs) and not publicly available. Given the lack of a reasonably large clinical QA test set for studying generalization, with the help of three clinical experts, we create 1287 QA pairs on a sampled set of MIMIC-III \citep{mimiciii} clinical notes, which have been reviewed and approved by PhysioNet.\footnote{{https://physionet.org/}. PhysioNet is a resource center with missions to conduct and catalyze for biomedical research, which offers free access to large collections of physiological and clinical data, such as MIMIC-III \citep{mimiciii}.}

\input{Table/Table_ch2_1}

\noindent\textbf{Annotation Process.} 
We sample 36 MIMIC-III clinical notes\footnote{When sampling MIMIC-III notes, we ensure that all the sampled clinical texts do not appear in emrQA, acknowledging that there is a small overlap between the two datasets.} as contexts. For each context, clinical experts can ask any questions as long as an answer can be extracted from the context. To save annotation effort, QA pairs generated by 9 QG models (i.e., all base QG models and their diversity-enhanced variants; see Section \ref{exp_setup}) are provided as references, and (nearly) duplicates are removed. Meanwhile,  clinical experts are highly encouraged to create new questions based on the given clinical text (which are marked as \textit{``human-generated"}). But if they do find that the machine-generated questions sound natural and match the provided answer, they can keep them (which are marked as \textit{``human-verified"}). After obtaining the annotated questions, we ask another clinical expert to do a final pass of the questions in order to further ensure the quality of the test set.  The final test set consists of 1287 questions (of which 975 are \textit{``human-verified"} and 312 are \textit{``human-generated"}).

In the following sections, we consider emrQA as the \textit{source} dataset and our annotated MIMIC-III QA dataset as the \textit{target} data. Detailed statistics of the two datasets are given in Table~\ref{tbl_ch2:dataset}.

\section{Framework}
\subsection{Overview of Our Framework}
\label{Overview_of_Our_Framework}
\begin{figure*}[t]
    \centering
    \includegraphics[width=0.95\linewidth]{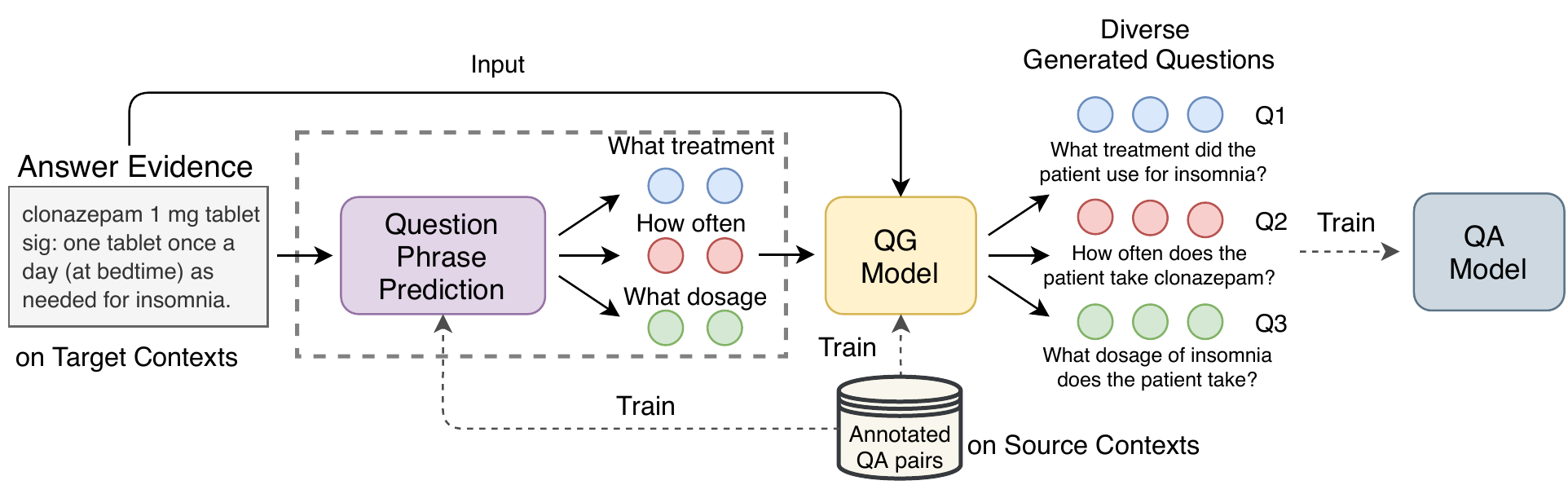}
    \caption[Illustration of our framework equipped with QPP.] {Illustration of our framework equipped with QPP: A key component is our question phrase prediction (QPP) module, which aims to generate diverse question phrases and can be ``plugged-and-played" with most existing QG models to diversify their generation.}
    \label{fig:framework}
\end{figure*}

We first give an overview of our framework without including any diversity booster.

To solve the proposed generalization challenge of clinical QA, inspired by recent work on question generation (QG) for QA in the open domain \citep{golub2017two,wang2019adversarial,ShakeriSZNNWNX20}, we implement an answer evidence extractor and a seq2seq-based neural QG model \cite[][NQG]{du2017learning} to synthesize QA pairs on target contexts. Specifically, given a document, we deploy a ClinicalBERT \citep{alsentzer2019publicly} model to extract a long text\footnote{Following \citet{pampari2018emrqa,yue2020CliniRC}, we focus on long text spans instead of short answers since the former often contain richer information, which is more useful to support clinical decision making.} span as an answer evidence. We formulate such span prediction problem as a \texttt{BIO} tagging task. After prediction, we develop some heuristic rules (e.g., removing/merging very short extracted evidences) to further improve the quality of the extracted evidences; more  details are listed in Appendix~\ref{apx:extract_evidence}.
Based on the extracted answer evidences, a seq2seq-based QG model can be used to generate questions. Both answer evidence extractor and QG model are trained on the source data and then used to synthesize QA pairs on target contexts, based on which a QA model can be trained.

\subsection{Preliminary Observation}
\begin{figure}[t]
    \centering
    \includegraphics[width=0.6\linewidth]{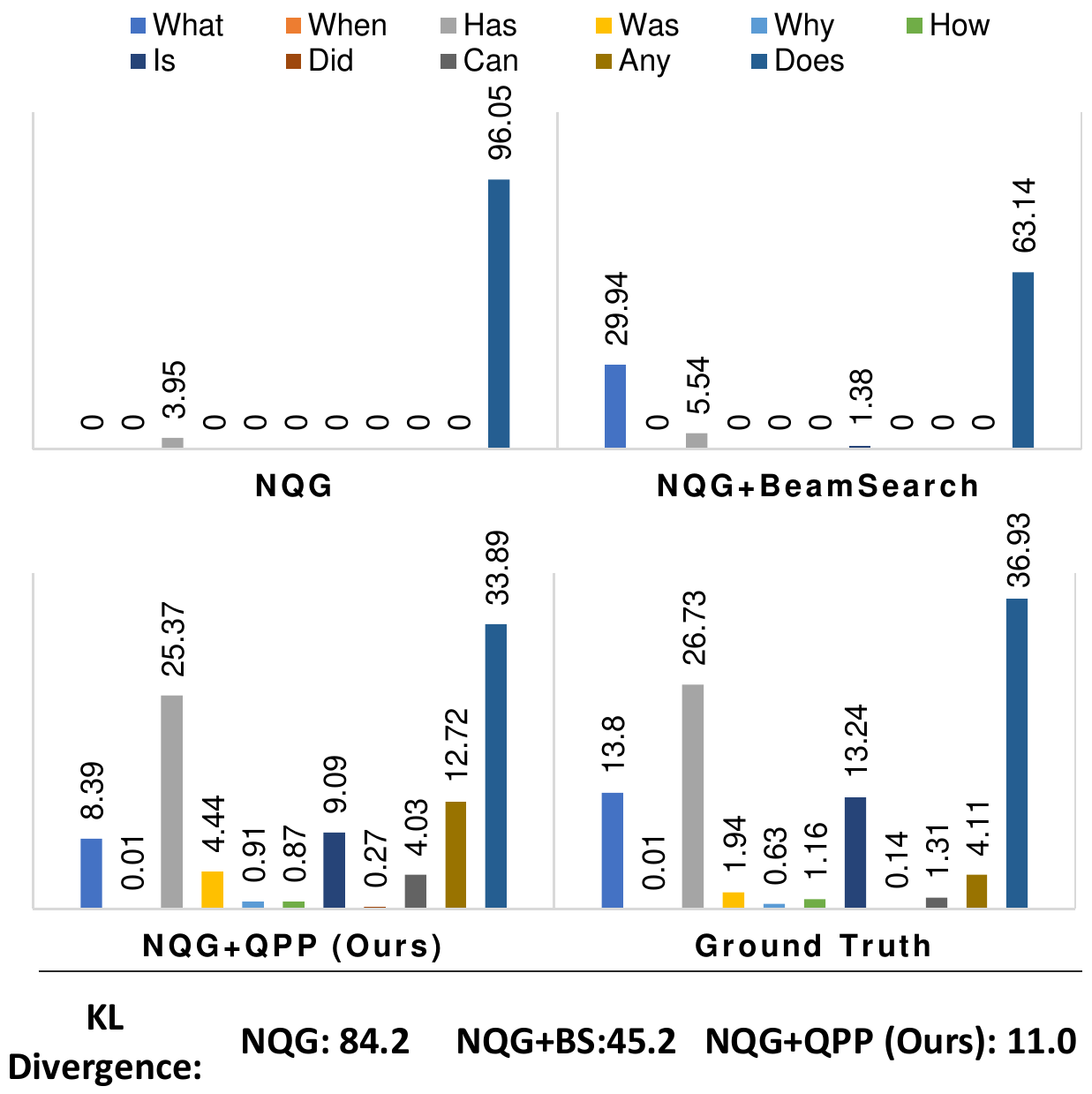}
    \caption[Distributions over types of questions generated by NQG models and the ground truth. ]{Distributions over types of questions generated by NQG models and the ground truth. BS: Beam Search; QPP: Question Phrase Prediction module.}
    \label{fig:ques_distibution}
\end{figure}

To our surprise, training on the synthesized target-context QA pairs does not yield an improvement of QA on the constructed MIMIC-III QA set. Specifically, F1 is 79.43 for the QA model trained on corpus synthesized by NQG (neural question generation) model, which is a little inferior to directly training the QA model on emrQA (79.99 F1). 
An outstanding characteristic we observe in the generated questions is the large bias of question types (e.g., most questions are ``Does" while there is few ``Why" and no ``How" question).
The distributions of question types are in Figure~\ref{fig:ques_distibution} (see top-left sub-plot).

\section{Diverse Question Generation for QA}
Given the observation above, we argue that the synthetic questions should be diverse so that they could serve as more useful training corpora. 

\subsection{Overview of Diverse Question Generation}

We investigate two kinds of approaches to diversify the generation. In the first decoding-based approach, we select the standard beam search as the representative since it is well studied and shows competitive performance in diversifying generations \citep{IppolitoKSKC19}. For the other kind (topic-guided approach), we propose a \textit{question phrase prediction (QPP)} module, which predicts a set of valid question phrases given an answer evidence (Figure \ref{fig:framework}). Then, conditioned on a question phrase sampled from the set predicted by the QPP, a QG model is utilized to complete the rest of the question.

\subsection{Question Phrase Prediction (QPP)}
We formulate the question phrase prediction task as a \emph{sequence prediction} problem and adopt a commonly used seq2seq model \citep{luong2015effective}. More formally, given an answer evidence $\textbf{a}$, QPP aims to predict a sequence of question phrases $\vec{s}=(s_{1}, ..., {s_{{|\vec{s}|}}})$(e.g., ``What treatment'' ($s_{1}$) $\rightarrow$ ``How often'' ($s_{2}$) $\rightarrow$ ``What dosage'' ($s_{3}$), with $|\vec{s}|=3$). 

\input{Table/Table_ch2_alg}

During training, we assume that the set of question phrases is arranged in a pre-defined order. Such orderings can be obtained with some heuristic methods, e.g., using a descending order based on question phrase frequency in the corpus\footnote{In emrQA, each answer evidence is tied with multiple questions, which allows the training for QPP.} (more details are in Appendix~\ref{apx:QPP_I}).
As such, we aim to minimize:
\vspace{-4pt}
\begin{equation}
\vspace{-4pt}
\label{eq:Loss_QPP}
    {L}_{QPP}=-\sum \log P (\vec{s}|\vec{a};\theta)
\end{equation}
where $\textbf{s}, \textbf{a}, \theta$ denote question phrase sequence, input answer evidence and all the parameters in QPP, respectively. Algorithm~\ref{alg:model} illustrates the pretraining and training procedure of our framework when equipped with our proposed QPP module.

In the inference stage, QPP can dynamically decide the number of question phrases for each answer evidence by predicting a special \texttt{[STOP]} type.
By decomposing QG into two steps (diversification followed by generation), the proposed QPP can increase the diversity in a more controllable way compared with decoding-based approach.

\section{Evaluation and Results}

\subsection{Experiment Setup}
\label{exp_setup}

\noindent\textbf{Base QG and QA models:} In our experiments, we adopt three base QG models: NQG \citep{du2017learning}, NQG++ \citep{zhou2017neural} and BERT-SQG \citep{chan2019recurrent}. For QA, we use two base models, DocReader \citep{chen2017reading} and ClinicalBERT \citep{alsentzer2019publicly}.

To investigate the effectiveness of diverse QG for QA, we consider the following variants of each base QG model: (1) Base Model: Inference with greedy search; (2) Base Model + Beam Search: Inference with Beam Search with the beam size at $K$ and keep top $K$ beams (we set $K=3$) (3) Base Model + QPP: Inference with greedy search for both QPP module and Base model. 

When training a QA model, we only use the synthetic data on the target contexts and do not combine the synthetic data with the source data since the combination does not help in our preliminary experiments.

\noindent\textbf{Evaluation Metrics:} For QG evaluation, we focus on evaluating both relevance and diversity. Following previous work \citep{du2017learning,zhang2018generating}, we use BLEU \citep{papineni2002bleu}, ROUGE-L \citep{lin-rouge} as well as METEOR \citep{lavie2009meteor} for relevance evaluation. Since the Beam Search and our QPP module enable QG models to generate multiple questions given an evidence, we report the top-1 relevance among the generated questions following \citet{cho2019mixture}. For diversity, we report Distinct \citep{li2016diversity} as well as Entropy \citep{zhang2018generating} scores. We calculate BLEU and the diversity measures based on $3$- and $4$-grams.

{For QA evaluation, we report exact match (EM) (the percentage of predictions that match the ground truth answers exactly) and F1 (the average overlap between the predictions and
ground truth answers) as in \citet{rajpurkar2016squad}.}

\input{Table/Table_ch2_3}

\subsection{Results}
Table~\ref{tbl_ch2:qa_results} summarizes the performance of two widely used QA models, DocReader \citep{chen2017reading} and ClinicalBERT \citep{alsentzer2019publicly}, on the MIMIC-III test set. The QA models are trained on different corpora, including the emrQA dataset as well as QA pairs generated by different models. 

We also evaluate QG models on the emrQA dataset (i.e., train and test QG solely on source domain). As can be seen from Table~\ref{tbl_ch2:qg_results}, the three selected base models (NQG, NQG++ and BERT-SQG) all achieve very promising relevance scores; however, they do not perform well with diversity scores.  The diversity of generated questions is boosted to some extent when the Beam Search is used since it can offer flexibility for QG models to explore more candidates when decoding. In comparison, the QPP module in our framework leads to the best results under both relevance and diversity evaluation. Particularly, it obtains $5\%$ absolute improvement in terms of Dist4 for each base model.
\input{Table/Table_ch2_2}
\section{Analysis}
\subsection{Quantitative Analysis}
\noindent\textbf{Analysis on QA Generalization:}
As expected, the corpora generated by diverse QG help the QA model perform consistently better than those generated by their respective base QG version as well as emrQA (Table~\ref{tbl_ch2:qa_results}).
Between the two diversity-boosting approaches, we observe that the QA model trained on the corpora by QPP-enhanced QG achieves the best performance.
Moreover, results on the human-generated portion are consistently better than those on human-verified. This is likely due to the fact that human-generated questions are more readable and sensible while human-verified ones are less natural (though the correctness is ensured). All these results indicate that improving the diversity of generated questions can help better train QA models on the new contexts and better address the generalization challenge.

\begin{figure}[t]
    \centering
    \includegraphics[width=\linewidth]{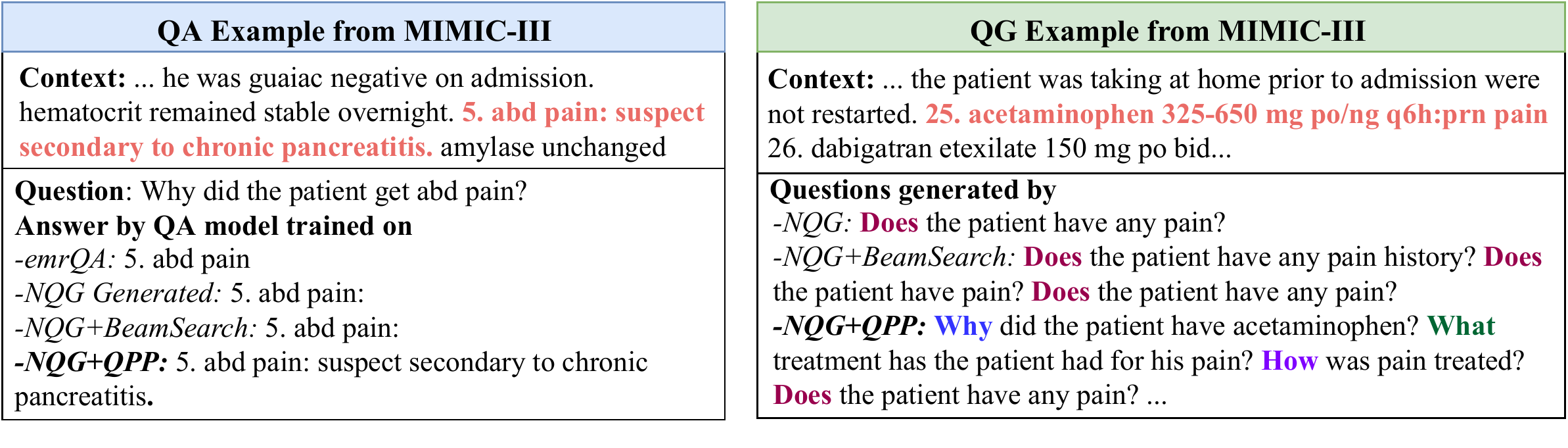}
    \caption[QA and QG examples.] {QA and QG examples. The red parts in contexts are ground-truth answer evidences. } 
    \label{fig:qualitative}
\end{figure}

\noindent\textbf{Analysis on QG diversity:}
Figure~\ref{fig:ques_distibution} shows the distribution over types of questions generated by NQG-based models (i.e., base model, base + beam search and base + QPP) and the ground truth on emrQA dataset. We observe that the Kullback–Leibler (KL) divergence between the distributions of generated questions and the ground truth is smaller after enabling diversity booster. The gap reaches the minimum when our QPP module is plugged in. It's worth noting that even some of the least frequent types of questions (e.g., ``How", ``Why") can be generated when our QPP module is turned on. These observations demonstrate diversity booster, especially our QPP module, can help generate diverse questions.

\subsection{Qualitative Analysis: Error Analysis}
In Figure \ref{fig:qualitative}, we first present a QA example and a QG example from MIMIC-III. In the QA example, this ``why" question can be correctly answered  by the QA model (DocReader) trained on the ``NQG+QPP" generated corpus while the QA models trained on other generated corpora fail. This is because the NQG model and ``NQG+BeamSearch" cannot generate any ``why" questions as shown in Figure \ref{fig:ques_distibution}. Thus QA models trained on such corpora cannot answer questions of less frequent types. Though the emrQA dataset contains diverse questions (including ``why" questions),  its contexts might be different from MIMIC-III in terms of topic, note structures, writing styles, etc.  So the model trained on emrQA struggles to answer some questions. In the QG example, the base model NQG can only generate one question. Though utilizing the Beam Search enables the model to explore multiple candidates, the generated questions are quite similar and are less likely to help improve QA. Enabling our QPP module helps generate diverse questions including ``Why", ``What", ``How", etc.

\section{Conclusion}
In this chapter, we systematically investigate the generalization challenge of clinical reading comprehension and construct a new test set on MIMIC-III clinical texts. After observing simply using QG for QA does not work, we explore the importance of generating \textit{diverse} questions. That is, we study two approaches for boosting question diversity, beam search and QPP. Particularly, our proposed QPP (question phrase prediction) module significantly improves the cross-domain generalizability of QA systems. Our comprehensive experiments allow for a better understanding of why diverse question generation can help QA on new clinical documents (i.e., target domain).

%% file: Table/Table_ch2_1.tex
\begin{table}[t]
\centering
\resizebox{0.6\linewidth}{!}{%
\begin{tabular}{ccc}
\hline
 (Question / Context) & emrQA & MIMIC-III \\ \hline
\# Train  & 781,857 / 337 & - / 337 \\
\# Dev  & 86,663 / 41 & 8,824 / 40 \\
\# Test & 98,994 / 42 & 1,287 / 36 \\
\# Total & 967,514 / 420 & - / 413 \\ \hline
for purpose of & \begin{tabular}[c]{@{}c@{}}QG \& QA \\ (\textit{source} domain)\end{tabular} & \begin{tabular}[c]{@{}c@{}}QA\\ (\textit{target} domain)\end{tabular} \\ \hline
\end{tabular}%
}
\caption[Statistics of the datasets.] {Statistics of the datasets. We synthesize a machine-generated dev set and ask human experts to annotate a test set for MIMIC-III. Details of dev set construction can be found in Setion~\ref{DevSetConstruction}.}
\label{tbl_ch2:dataset}
\end{table}

%% file: Table/Table_ch2_alg.tex
\newcommand\Algphase[1]{%
\vspace*{-.7\baselineskip}\Statex\hspace*{\dimexpr-\algorithmicindent-0pt\relax}\rule{1.08\linewidth}{0.6pt}%
\vspace{-4pt}
\Statex\hspace*{-\algorithmicindent}\textbf{#1}%
\vspace*{-.7 \baselineskip}\Statex\hspace*{\dimexpr-\algorithmicindent-0pt\relax}\rule{1.08\linewidth}{0.6pt}%
}
\setlength{\textfloatsep}{2pt}%

\begin{algorithm}[t]
 \caption{Training procedure of our framework equipped with QPP.}
 \label{alg:model}
 \textbf{Input}: labeled \textit{source} data $\{(P_S,A_S,Q_S)\}$, unlabeled \textit{target} data $\{P_T\}$ \\
 \textbf{Output}: Generated QA pairs $\{(A'_T,Q'_T)\}$  on \textit{target} contexts; An optimized QA model for answering questions on target contexts
 \begin{algorithmic}[1]
 \Algphase{Pretraining Stage}
\State Train \textit{Answer Evidence Extractor} based on the \textit{source} data $\{(P_S,A_S)\}$

\State Obtain question phrase data $Y_S$ from $Q_S$ and  train \textit{Question Phrase Prediction} module on the \textit{source} data $\{(A_S, Y_S)\}$

\State Train a \textit{QPP-enhanced QG} model on the \textit{source} data $\{(A_S, Y_S,Q_S)\}$ 

\Algphase{Training Stage}

\State Use \textit{AEE} to extract potential answer evidences $\{A'_T\}$ on the \textit{target} contexts $\{P_S\}$

\State Use \textit{QPP} to predict potential question phrases set $\{Y'_T\}$ on $\{A'_T\}$ 

\State Use \textit{QPP-enhanced QG} to generate diverse questions $\{Q'_T\}$ based on  $\{(A'_T,Y'_T)\}$

\State Train a \textit{QA} model on synthetic \textit{target} data $\{(P_T, A'_T,Q'_T)\}$ 
\end{algorithmic} 
\end{algorithm}

%% file: Table/Table_ch2_3.tex
\begin{table*}[t]
\centering
\resizebox{\linewidth}{!}{%
\begin{tabular}{l|cccccc|cccccc}
\hline
\multirow{4}{*}{\textbf{QA Datasets}} & \multicolumn{6}{c|}{\textbf{DocReader} \citep{chen2017reading}} & \multicolumn{6}{c}{\textbf{ClinicalBERT} \citep{alsentzer2019publicly}} \\ \cline{2-13} 
 & \multicolumn{2}{c|}{\begin{tabular}[c]{@{}c@{}} \textbf{Human}\\  \textbf{Verified} \end{tabular}} & \multicolumn{2}{c|}{{\begin{tabular}[c]{@{}c@{}} \textbf{Human}\\  \textbf{Generated} \end{tabular}} } & \multicolumn{2}{c|}{{\begin{tabular}[c]{@{}c@{}} \textbf{Overall}\\  \textbf{Test} \end{tabular}} } & \multicolumn{2}{c|}{\begin{tabular}[c]{@{}c@{}} \textbf{Human}\\  \textbf{Verified} \end{tabular}} & \multicolumn{2}{c|}{{\begin{tabular}[c]{@{}c@{}} \textbf{Human}\\  \textbf{Generated} \end{tabular}} } & \multicolumn{2}{c}{{\begin{tabular}[c]{@{}c@{}} \textbf{Overall}\\  \textbf{Test} \end{tabular}} } \\ \cline{2-13} 
 & \textbf{EM} & \textbf{F1} & \textbf{EM} & \textbf{F1} & \textbf{EM} & \textbf{F1} & \textbf{EM} & \textbf{F1} & \textbf{EM} & \textbf{F1} & \textbf{EM} & \textbf{F1}  \\ \hline
 {\begin{tabular}[l]{@{}l@{}} emrQA \citep{pampari2018emrqa} \end{tabular}} 
  & 61.44 & 78.82 & 69.87 & 83.66 & 63.48 & 79.99 & 61.23 & 78.56 & 69.23 & 82.83 & 63.17 & 79.59  \\ \hline
 {\begin{tabular}[c]{@{}c@{}} NQG \citep{du2017learning}\\      \end{tabular}} 
 & 64.71 & 79.36 & 66.99 & 79.67 & 65.26 & 79.43 & 59.49 & 76.68 & 67.3 & 82.59 & 61.38 & 78.11\\
+ BeamSearch & 67.07 & 81.21 & 71.15 & 83.07 & 68.07 & 81.66 & 63.17 & 79.17 & 68.91 & 84.26 & 64.56 & 80.4 \\
\textbf{+ QPP (Ours)} & \textbf{68.82} & \textbf{82.89} & \textbf{74.68} & \textbf{85.18} & \textbf{70.09} & \textbf{83.44} & \textbf{63.79} & \textbf{79.56} & \textbf{69.23} & \textbf{84.33} & \textbf{65.11} & \textbf{80.72}\\ \hline
 {\begin{tabular}[c]{@{}c@{}} NQG++ \citep{zhou2017neural}  \end{tabular}} & 65.94 & 78.71 & 66.34 & 81.34 & 66.04 & 79.35 & 59.59 & 75.85 & 65.06 & 80.11 & 60.92 & 76.88\\
+ BeamSearch & 68.10 & 80.09 & 72.11 & 84.56 & 69.07 & 81.17 & 64.61 & 80.30 & 68.26 & 83.70 & 65.50 & 81.12  \\
\textbf{+ QPP (Ours)} & \textbf{70.05} & \textbf{83.47} & \textbf{74.36} & \textbf{85.92} & \textbf{71.10} & \textbf{84.06} & \textbf{65.33} & \textbf{80.64} & \textbf{70.83} & \textbf{85.76} & \textbf{66.67} & \textbf{81.88} \\ \hline
\defcitealias{chan2019recurrent}{[Chan et al., 2019]}
{\begin{tabular}[c]{@{}c@{}} BERT-SQG \citetalias{chan2019recurrent}  \end{tabular}}
 & 66.05 & 79.64 & 70.19 & 81.47 & 67.05 & 80.08 & 59.59 & 78.04 & 65.06 & 82.20 & 60.92 & 79.05 \\
+ BeamSearch & 68.71 & 81.98 & 73.71 & 84.44 & 69.93 & 82.58 & 61.94 & 79.02 & 67.31 & 82.54 & 63.25 & 79.88\\
\textbf{+ QPP (Ours)} & \textbf{70.77} & \textbf{83.60} & \textbf{74.36} & \textbf{85.53} & \textbf{71.64} & \textbf{84.07}  & \textbf{64.21} & \textbf{80.53} & \textbf{69.23} & \textbf{85.38} & \textbf{65.43} & \textbf{81.71}\\ \hline
\end{tabular}%
}
\caption[QA performance on MIMIC-III test set.] {QA performance on MIMIC-III test set. emrQA is also included as a baseline dataset to illustrate that the generated diverse questions on MIMIC-III are useful to improve the QA model performance on new contexts.}
\label{tbl_ch2:qa_results}
\end{table*}

%% file: Table/Table_ch2_2.tex
\begin{table*}[t]
\centering
\resizebox{\linewidth}{!}{%
\begin{tabular}{lcccc|cccc}
\hline
\multirow{2}{*}{\textbf{Models}} & \multicolumn{4}{c|}{\textbf{Relevance}} & \multicolumn{4}{c}{\textbf{Diversity}} \\ \cline{2-9} 
 & \textbf{BLEU3} & \textbf{BLEU4} & \textbf{MR} & \textbf{RG} & \textbf{Dist3} & \textbf{Dist4} & \textbf{Ent3} & \textbf{Ent4} \\ \hline 
NQG \citep{du2017learning} & 91.45 & 90.11 & 60.70 & 94.62 & 0.233 & 0.282 & 4.473 & 4.738 \\
+ BeamSearch & 94.33 & 93.42 & 62.08 & 95.56 & 0.569 & 0.775 & 5.406 & 5.812 \\
\textbf{+ QPP (Ours)} & \textbf{96.82} & \textbf{96.33} & \textbf{64.38} & \textbf{97.49} & \textbf{3.177} & \textbf{5.289} & \textbf{7.100} & \textbf{7.777} \\ \hline
NQG++ \citep{zhou2017neural} & 97.11 & 96.65 & 71.57 & 97.86 & 0.229 & 0.275 & 4.419 & 4.648 \\
+ BeamSearch & 98.35 & 98.07 & 72.98 & 98.55 & 0.618 & 0.848 & 5.497 & 5.953 \\
\textbf{+ QPP (Ours)} & {\textbf{99.15}} & {\textbf{99.03}} & {\textbf{74.01}} & {\textbf{99.11}} & \textbf{3.183} &  \textbf{5.293} & \textbf{7.111} & \textbf{7.798} \\ \hline
BERT-SQG \citep{chan2019recurrent} & 89.07 & 87.99 & 65.25 & 94.91 & 0.228 & 0.276 & 4.594 & 4.849 \\
+ BeamSearch &95.45   &94.84  &66.39  &96.22  &0.510  &0.713  &5.522  &6.015  \\
\textbf{+ QPP (Ours)} & \textbf{96.54} & \textbf{96.19} & \textbf{67.51} & \textbf{97.42} & {\textbf{3.344}} &{\textbf{5.332}} & {\textbf{7.173}} & {\textbf{7.816}} \\ \hline
\end{tabular}%
}
\caption[Automatic evaluation of the generated questions on emrQA dataset.] {Automatic evaluation of the generated questions on emrQA dataset. For each base model, the best performing variant is in \textbf{bold}. RG: ROUGE-L, MR: METEOR, Dist: Distinct, Ent: Entropy.}
\label{tbl_ch2:qg_results}
\end{table*}

%% file: ch4.NLI.tex
\chapter{Natural Language Inference
\label{NLI}}

\section{Introduction}

\input{Table/Table_ch4_1}

Natural language inference (NLI)\footnote{In this thesis, we use ``textual entailment'' and ``Natural Language Inference'' or ``NLI'' interchangeably.} is the problem of determining whether a natural
language hypothesis \textit{h} can be inferred (or entailed) from a natural language premise \textit{p} \citep[i.a.,][]{DaganGM05,MacCartneyM09}. Conventionally, people only examine items that are suitable for systematic inferences (i.e., items for which people consistently agree on the NLI label).

However, \citet{PavlickK19} observed inherent disagreements among annotators in several NLI datasets (e.g., SNLI \citep{bowman-etal-2015-large}), which cannot be smoothed out by hiring more people. They pointed out that to achieve robust NLU, we need to be able to tease apart systematic inferences (i.e., items for which most people agree on the annotations) from items inherently leading to disagreement. The last example in Table~\ref{tbl_ch4:examples} is a typical disagreement item: some annotators consider it to be an entailment (3 or 2), while others view it as a contradiction (-3). Clearly, the annotators have two different interpretations on the complement clause “If she’d said Carolyn had borrowed a book from Clare and wanted to return it”. Moreover, a common practice in the literature to generate an inference label from annotations is to take the average \citep[i.a.,][]{PavlickC16}. In this case, it would be ``Neutral'', but such label is not accurately capturing the distribution. Alternatively, some work simply ignored the ``Disagreement'' portion but only studied systematic inferences items \citep{JiangM19ACL,JiangM19, Raffel2019}.

\citet{Kenyon-Dean18} also pointed out in sentiment analysis task, when performing real-time sentiment classification, an automated system cannot know a priori whether the data sample is inherently non-ambiguous. Here, in line with what \citet{Kenyon-Dean18} suggested for sentiment analysis, we propose a finer-grained labeling for NLI: teasing disagreement items, labeled ``Disagreement", from systematic inferences, which can be ``Contradiction", "Neutral" or ``Entailment".  As such, in order to achieve robust NLU in NLI task, the developed models should be able to identify inherent disagreement items when possible and carry out systematic inferences on non-disagreement items.

To this end, we propose Artificial Annotators (AAs), an ensemble of BERT models \citep{DevlinCLT19}, which simulate the uncertainty in the annotation process by capturing modes in annotations. That is, we expect to utilize simulated modes of annotations to enhance finer-grained NLI label prediction. Our results, on the CommitmentBank, show that AAs perform statistically significantly better than all baselines (including BERT baselines) by a large margin in terms of both F1 and accuracy. We also show that AAs manage to learn linguistic patterns and context-dependent reasoning.

\section{Inherently Ambiguous Items in CB}

We start with the introduction to the dataset used in this chapter, CommitmentBank \citep{deMarneffe19}, and then move on to how we determine ambiguous items and systematic inference items.

The CommitmentBank (CB) is a corpus of 1,200 naturally occurring discourses originally collected from news articles, fiction and dialogues. Each discourse consists of up to 2 prior context sentences and 1 target sentence with a clause-embedding predicate under 4 embedding environments (negation, modal, question or antecedent of conditional). Annotators judged the extent to which the speaker/author of the sentences is committed to the truth of the content of the embedded clause (CC), responding on a Likert scale from +3 to -3, labeled at 3 points (+3/speaker is certain the CC is true, 0/speaker is not certain whether the CC is true or false, -3/speaker is certain the CC is false).
Following \citet{JiangM19}, we recast CB by taking the context and target as the premise and the embedded clause in the target as the hypothesis.

Common NLI benchmark datasets are SNLI \citep{bowman-etal-2015-large} and MultiNLI \citep{williams-etal-2018-broad}, but these datasets have only one annotation per item in the training set. CB has at least 8 annotations per item, which permits to identify items on which annotators disagree. \citet{JiangM19} discarded items if less than 80\% of the annotations are
within one of the following three ranges: [1,3] Entailment, 0 Neutral, [-3,-1] Contradiction. The gold label for example 3 in Table~\ref{tbl_ch4:examples} would thus be ``Disagreement''. However, this seems a bit too stringent, given that ~70\% of the annotators all agree on the 0 label and there is only one annotation towards the extreme. Likewise, for example 5, most annotators chose a negative score and the item might therefore be better labeled as ``Contradiction'' rather than ``Disagreement''. To decide on the \textbf{finer-grained NLI labels}, we therefore also took variance and mean into account, as follows:\footnote{Compared with the labeling scheme in \citet{JiangM19}, our labeling scheme results in 59 fewer Disagreement items, 48 of which are labeled as Neutral.} 

\begin{itemize}[leftmargin=2.2em,noitemsep,topsep=0pt,parsep=0pt,partopsep=0pt]
\item \textbf{Entailment:} 80\% of annotations fall in the range [1,3] OR the annotation variance $\leq$ 1 and the annotation mean $>$ 1.
\item \textbf{Neutral:} 80\% of annotations is 0 OR the annotation variance $\leq$ 1 and the absolute mean of annotations is bound within 0.5.
\item \textbf{Contradiction:} 80\% of annotations fall in the range [-3, -1] OR the annotation variance $\leq$ 1 and the annotation mean $<$ -1.
\item \textbf{Disagreement:} Items which do not fall in any of the three categories above.
\end{itemize}

\input{Table/Table_ch4_7}

We randomly split CB into train/dev/test sets in a 7:1:2 ratio.\footnote{We don't follow the SuperGLUE splits \citep{WangPNSMHLB19} as they do not include disagreement items.} Table~\ref{tbl_ch4:statistics} gives splits' basic statistics.

\section{Linguistic Rules}
\label{ling_rule}

Our developed linguistic rules are inspired by and adapted from \citet{JiangM19} to explicitly include the most discriminating expressions for disagreement items. We utilize three linguistic features which are provided in CB: entailment-canceling environment (negation, modal, question, antecedent of conditional), matrix verb and its subject person.

\begin{itemize}[leftmargin=2.8em,noitemsep,topsep=0pt,parsep=0pt,partopsep=0pt]
    \item[1.] Items under conditional are disagreement.
    \item[2.] Items under question and with second person are neutral.
    \item[3.] Items under question and with non-second person are disagreement.
    \item[4.] Items of the form "I don't know/think/believe" are contradiction (i.e., negRaising structure).
    \item[5.] Items with factive verbs are entailment.
    \item[6.] Items under negation and with non-factive verbs are disagreement.
    \item[7.] Items under modal and with non-third person are entailment.
\end{itemize}

When this policy is executed, there are two additional auxiliary rules: Items not falling in any group above are assigned a disagreement label as it is the dominant class in CB; For items satisfying more than one rule, the label will be determined by the higher-ranked rule (a smaller number indicates a higher rank).
Note that the rules above also reveal the most discriminating expressions for each class.

\section{Artificial Annotators}
We aim at finding an effective way to tease items leading to systematic inferences apart from items leading to disagreement. 
As pointed out by \citet{CalmaS17}, annotated labels are subject to uncertainty. Annotations are indeed influenced by several factors: workers' past experience and concentration level, cognition complexities of items, etc. They proposed to simulate the annotation process in an active learning paradigm to make use of the annotations that contribute to uncertainty. Likewise, for NLI, \citet{Gantt2020} observed that directly training on raw annotations using annotator identifier improves performance. Essentially, \citet{Gantt2020} used a mixed-effect model to learn a mapping from an item and the associated annotator identifier to a NLI label. However, annotator identifiers are not always accessible, especially in many datasets that have been there for a while. Thus, we decide to simulate the annotation process instead of learning from real identifiers.

As shown by \citet{PavlickK19}, if annotations of an item follow unimodal distributions, then it is suitable to use aggregation (i.e., take an average) to obtain a inference label; but such an aggregation is not appropriate when annotations follow multi-modal distributions. Without loss of generality, we assume that items are associated with n-modal distributions, where n $\geq$ 1. Usually, systematic inference items are tied to unimodal annotations while disagreement items are tied to multi-modal annotations. We, thus, introduce the notion of Artificial Annotators (AAs), where each individual ``annotator'' learns to model one mode. 

\subsection{Architecture} AAs is an ensemble of $n$ BERT models \citep{DevlinCLT19} with a primary goal of finer-grained NLI label prediction. $n$ is determined to be 3 as there are up to 3 relationships between premise and hypothesis, excluding the disagreement class. Within AAs, each BERT is trained for an auxiliary systematic inference task which is to predict entailment/neutral/contradiction based on a respective subset of annotations. The subsets of annotations for the three BERT are mutually exclusive. 

\begin{figure}
  \centering
  \includegraphics[width=0.8\textwidth]{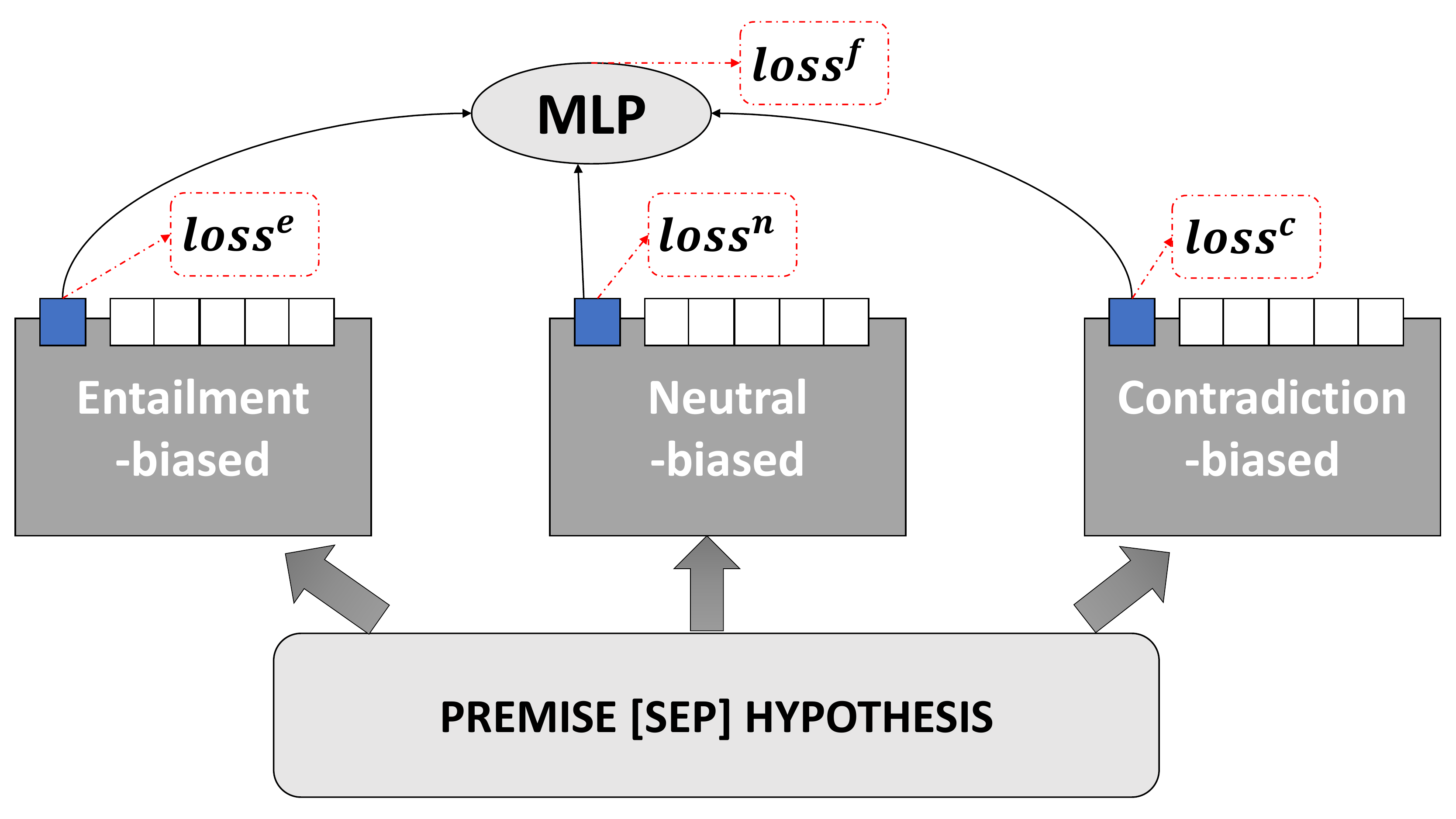}
    \caption{Artificial Annotators (AAs) setup.}
  \label{fig_ch4:model}
\end{figure}

A high-level overview of AAs is shown in Figure~\ref{fig_ch4:model}. Intuitively, each BERT separately predicts a systematic inference label, each of which represents a mode\footnote{It's possible that three modes collapse to (almost) a point.} of the annotations. The representations of these three labels are further aggregated as augmented information to enhance final fine-grained NLI label prediction (see Eq.~\ref{ch4_MLP}).  

If we view the AAs as a committee of three members, our architecture is reminiscent of the Query by Committee (QBC) \citep{SeungOS92}, an effective approach for active learning paradigm. The essence of QBC is to select unlabeled data for labeling on which disagreement among committee members (i.e., learners pre-trained on the same labeled data) occurs. The selected data will be labeled by an oracle (e.g., domain experts) and then used to further train the learners. Likewise, in our approach, each AA votes for an item independently. However, the purpose is to detect disagreements instead of using disagreements as a measure to select items for further annotations. Moreover, in our AAs, the three members are trained on three disjoint annotation partitions for each item (see Section~\ref{ch4_training}).

\subsection{Training} 
\label{ch4_training}
We first sort the annotations in descending order for each item and divide them into three partitions.\footnote{For example, if there are 8 annotations for a given item, the annotations are divided into partitions of size 3, 2 and 3.} For each partition, we generate an auxiliary label derived from the annotation mean. If the mean is greater/smaller than +0.5/-0.5, then it's entailment/contradiction; otherwise, it's neutral. The first BERT model is always enforced to predict the auxiliary label of the first partition to simulate an entailment-biased annotator. Likewise, the second and third BERT models are trained to simulate neutral-biased and contradiction-biased annotators.

Each BERT produces a pooled representation for the \texttt{[CLS]} token. The three  representations are passed through a multi-layer perceptron (MLP) to obtain the finer-grained NLI label:
\vspace{-0pt}
\begin{equation}
    P(y|\mathbf{x}) = \operatorname*{softmax}(\mathbf{W_s}\tanh(\mathbf{W_t[e;n;c]}))
\label{ch4_MLP}
\vspace{-0pt}
\end{equation}
with $\mathbf{[e;n;c]}$ being the concatenation of three learned representations out of \textbf{e}ntailment-biased, \textbf{n}eutral-biased and \textbf{c}ontradiction-biased BERT models.  $\mathbf{W_s}$ and $\mathbf{W_t}$ are parameters to be learned.

The overall loss is defined as the weighted sums of four cross-entropy losses:
\begin{equation}
\small
    loss = r*loss^f + \frac{1-r}{3}(loss^e + loss^n + loss^c)
\label{ch4_loss}
\vspace{-0pt}
\end{equation}
where $r \in [0, 1]$ controls the primary finer-grained NLI label prediction task loss ratio.

\section{Evaluation and Results}

\noindent \textbf{Evaluation Setting:} We include five baselines to compare with:
\begin{itemize}[leftmargin=1em,noitemsep,topsep=0pt,parsep=0pt,partopsep=0pt]
\item \textbf{``Always 0''}: Always predict Disagreement.
\item \textbf{CBOW} (Continuous Bags of Words): Each item is represented as the average of its tokens' GLOVE vectors \citep{PenningtonSM14}.
\item \textbf{Heuristic baseline}: Linguistics-driven rules (detailed out in chapter~\ref{ling_rule}), adapted from \citet{JiangM19}; e.g., conditional environment discriminates for disagreement items.  
\item \textbf{Vanilla BERT}: \citep{DevlinCLT19} Straightforwardly predict among 4 finer-grained NLI labels.
\item \textbf{Joint BERT}: Two BERT models are jointly trained, each of which has a different speciality. The first one (2-way) identifies whether a sentence pair is a disagreement item. If not, this item is fed into the second BERT (3-way) which carries out systematic inference. 
\end{itemize}

For all baselines involving BERT, we follow the standard practice of concatenating the premise and the hypothesis with \texttt{[SEP]}.  

\input{Table/Table_ch4_2}

\noindent \textbf{Results:} Table~\ref{tbl_ch4:result} gives the accuracy and F1 for each baseline and AAs, on the CB dev and test sets.  We run each model 10 times, and report the average. Also, Our AAs achieve the lowest standard deviations on test set items compared to BERT-based models, indicating that it is more stable and potentially more robust to wild environments.

\section{Analysis}

\subsection{Empirical Results Analysis}
CBOW is essentially the same as the ``Always 0'' baseline as it keeps predicting Disagreement regardless of the input. The Heuristic baseline achieves competitive performance on the dev set, though it has a significantly worse result on the test set. Not surprisingly, both BERT-based baselines outperform the Heuristic on the test set: fine-tuning BERT often lead to better performance, including for NLI \citep{PetersRS19,McCoyPL19}. These observations are consistent with \citet{JiangM19} who observed a similar trend, though only on systematic inferences. Our proposed AAs perform consistently better than all baselines, and statistically significantly better on the test set (t-test, p $\leq$ 0.01).

\input{Table/Table_ch4_4}

\input{Table/Table_ch4_8}

 Table~\ref{tbl_ch4:result} also gives F1 for each class on the test set. AAs outperform all BERT-based models under all classes. However, compared with the Heuristic, AAs show an inferior result on ``Neutral'' items mainly due to the lack of ``Neutral'' training data. The first 4 examples in Table~\ref{tbl_ch4:qualitative} show examples for which AAs make the correct prediction while other baselines might not. The confusion matrix in Table~\ref{tbl_ch4:error_analysis} shows that the majority ($\sim$60\%) of errors come from wrongly predicting a systematic inference item as a disagreement item. In 91\% of such errors, AAs predict that there is more than one mode for the annotation (i.e., the three labels predicted by individual ``annotators'' in AAs are not unanimous), as in example 5 in Table~{~\ref{tbl_ch4:qualitative}}. AAs are thus predicting more modes than necessary when the annotation is actually following a uni-modal distribution. On the contrary, when the item is supposed to be a disagreement item but is missed by AAs (as in example 6 and 7 in Table~\ref{tbl_ch4:qualitative}), AAs mistakenly predict that there is only one mode in the annotations 78\% of the time. It thus seems that a method which captures accurately the number of modes in the annotation distribution would lead to a better model.

\subsection{Linguistic Construction Analysis}
We also examine the model performance for different linguistic constructions to investigate whether the model learns some of the linguistic patterns present in the Heuristic baseline. The Heuristic rules are strongly tied to the embedding environments. Another construction used is one which can lead to ``neg-raising'' reading, where a negation in the matrix clause is interpreted as negating the content of the complement, as in example 3 (Table~\ref{tbl_ch4:qualitative}) 
where \textit{I \textbf{do not think} they have seen a really high improvement} is interpreted as \textit{I think they \textbf{did not see} a really high improvement}. ``Neg-raising'' readings often occur with \textit{know}, \textit{believe} or \textit{think} in the first person under negation. There are 85 such items in the test set: 41 contradictions (thus neg-raising items), 39 disagreements and 5 entailments. Context determines whether a neg-raising inference is triggered \citep{An1908}.

\input{Table/Table_ch4_5}

Table~\ref{tbl_ch4:additional} gives F1 scores for the Heuristic, BERT models and AAs for items under the different embedding environments and potential neg-raising items in the test set. Though AAs achieve the best overall results, it suffers under conditional and question environments, as the corresponding training data is scarce (9.04\% and 14.17\%, respectively). 
The Heuristic baseline always assigns contradiction to the ``I don't know/believe/think'' items, thus capturing all 41 neg-raising items but missing disagreements and entailments. BERT, a SOTA NLP model, is not great at capturing such items either: 71.64 F1 on contradiction vs.\ 52.84 on the others (Vanilla BERT); 71.69 F1 vs.\ 56.16 (Joint BERT). Our AAs capture neg-raising items better with 77.26 F1 vs.\ 59.38, showing an ability to carry out context-dependent inference on top of the learned linguistic patterns. Table~\ref{tbl_ch4:linguistics}, comparing performance on test items correctly predicted by the linguistic rules vs.\ items for which context-dependent reasoning is necessary, confirms this: AAs outperform the BERT baselines in both categories.

\input{Table/Table_ch4_6}

\section{Conclusion}

In this chapter, we introduced finer-grained natural language inference. This task aims at teasing systematic inferences from inherent disagreements. The inherent disagreement items are challenging for NLU models to handle, rarely studied in past NLI work. We show that our proposed AAs, which simulate the uncertainty in annotation process by capturing the modes in annotations, perform statistically significantly better than all baselines. However the performance obtained ($\sim$66\%) is still far from achieving truly robust NLU, leaving room for improvement.

%% file: Table/Table_ch4_1.tex
\begin{table}
\resizebox{\linewidth}{!}{%
\begin{tabular}{lp{1.41\linewidth}l}
\toprule
1 & \textit{Premise:} Some of them, like for instance the farm in Connecticut, are quite small. If I like a place I buy it. I guess you could say it's a hobby. \\ 
& \textit{Hypothesis:} buying places is a hobby. \\
& \texttt{Entailment (Entailment)} [3, 3, 2, 2, 2, 2, 1, 1] \\ \midrule
2 & \textit{Premise:} ``I hope you are settling down   and the cat is well.'' This was a lie. She did not hope the cat was well. \\ 
& \textit{Hypothesis:}  the cat was well. \\
& \texttt{Neutral (Neutral)} [0, 0, 0, 0, 0, 0, 0, 0, -3] \\ \midrule
3 & \textit{Premise:} ``All right, so it wasn't the bottle by the bed. What was it, then?'' Cobalt shook his head which might have meant he didn't know or might have been admonishment for Oliver who was still holding the bottle of wine.\\ 
& \textit{Hypothesis:}  Cobalt didn't know. \\
& \texttt{Neutral (Disagreement)} [1, 0, 0, 0, 0, 0, 0, -2] \\ \midrule
4 & \textit{Premise:} A: No, it doesn't. B: And, of course, your court system when you get into the appeals, I don't believe criminal is in a court by itself. \\
& \textit{Hypothesis:} criminal is in a court by itself. \\
& \texttt{Contradiction (Contradiction)} [-1, -1, -2, -2, -2, -2, -2, -3] \\ \midrule
5 & \textit{Premise:} A: The last one I saw was Dances With The Wolves. B: Yeah, we talked about that one too. And he said he didn't think it should have gotten all those awards. \\
& \textit{Hypothesis:} Dances with the Wolves should have gotten all those awards. \\
& \texttt{Contradiction (Disagreement)} [0, 0, -1, -1, -2, -2, -2, -3] \\ \midrule
6 & \textit{Premise:} Meg realized she'd been a complete fool. She could have said it differently. If she'd said Carolyn had borrowed a book from Clare and wanted to return it they 'd have given her the address.\\
& \textit{Hypothesis:} Carolyn had borrowed a book from Clare. \\
& \texttt{Disagreement (Disagreement)} [3, 3, 3, 2, 0, -3, -3, -3] \\ \bottomrule
\end{tabular}
}
\caption[Examples from CommitmentBank.]{Examples from CommitmentBank, with finer-grained NLI labels. The labels in parentheses come from \citet{JiangM19}. Scores in brackets are the raw human annotations.}
\label{tbl_ch4:examples}
\end{table}

%% file: Table/Table_ch4_7.tex
\begin{table}
\centering
\resizebox{0.6\linewidth}{!}{%
\begin{tabular}{lrrrrr}
\toprule
 & Entailment & Neutral & Contradiction & Disagreement & Total \\\midrule 
Train & 177 & 57 & 196 & 410 & 840 \\
Dev & 23 & 9 & 22 & 66 & 120 \\
Test & 58 & 19 & 54 & 109 & 240 \\ \midrule
Total & 258 & 85 & 272 & 585 & 1,200 \\ \bottomrule
\end{tabular}
}
\caption{Number of items in each class in train/dev/test.}
\label{tbl_ch4:statistics}
\end{table}

%% file: Table/Table_ch4_2.tex
\begin{table*}
\centering
\resizebox{0.8\linewidth}{!}{%
\begin{tabular}{lrrrrp{0.2cm}rrrr}
\toprule
\multicolumn{1}{c}{\multirow{2}{*}{}} & \multicolumn{2}{c}{Dev} & \multicolumn{7}{c}{Test} \\ 
\cmidrule(lr){2-3}
\cmidrule(lr){4-10}
\multicolumn{1}{c}{} & Acc. & F1 & Acc. & F1 && Entail & Neutral & Contradict & Disagree \\ \midrule
Always 0 & 55.00 & 39.03 & 45.42 & 28.37  && 0.00 & 0.00 & 0.00 & 62.46 \\
CBOW & 55.25 & 40.54 & 45.09 & 28.37 && 0.00 & 0.00 & 0.69 & 62.17 \\
Heuristic & 65.00  & 62.08  & 54.17 & 50.60 && 22.54 & \textbf{52.94} & 64.46 & 58.20\\
Vanilla BERT & 63.71 & 63.54 & 62.50 & 61.93 && 59.26 & 49.64 & 69.09 & 61.93  \\ 
Joint BERT & 64.47 & 64.28 & 62.61 & 62.07 && 59.77 & 47.27 & 67.36 & 63.21 \\ \midrule
AAs (\texttt{ours}) & \textbf{65.15} & \textbf{64.41} & \textbf{65.60*} & \textbf{64.97*} && \textbf{61.07} & 51.27 & \textbf{70.89} & \textbf{66.49*}\\ \bottomrule
\end{tabular}
}
\caption[Baselines and AAs overall performance on CB dev and test sets, and F1 scores of each class on the test set.]{Baselines and AAs overall performance on CB dev and test sets, and F1 scores of each class on the test set (average of 10 runs). * indicates statistically significant difference (t-test, p $\leq$ 0.01).}
\label{tbl_ch4:result}
\end{table*}

%% file: Table/Table_ch4_4.tex
\begin{table}
\resizebox{\linewidth}{!}{%
\begin{tabular}{lp{1.4\linewidth}l}
\toprule
\toprule
1 & \textit{Premise:} B: Yeah, it is. A: For instance, B: I'm a historian, and my father had kept them, I think, since nineteen twenty-seven uh, but he burned the ones from twenty-seven to fi-, A: My goodness. B: I could not believe he did that,\\
& \textit{Hypothesis:} his father burned the ones from twenty-seven \\
& Heuristics: \texttt{C} \hspace{0.2cm} V. BERT: \texttt{D} \hspace{0.2cm} J. BERT: \texttt{E} \hspace{0.2cm}  AAs: \texttt{E} \{\texttt{E}, \texttt{E}, \texttt{E}\}  \\ 
&  Gold: \texttt{E} [3, 3, 3, 3, 3, 2, 2, -1]\\ \midrule
2 & \textit{Premise:} `She was about to tell him that was his own stupid fault and that she wasn't here to wait on him - particularly since he had proved to be so inhospitable. But she bit back the words. Perhaps if she made herself useful he might decide she could stay - for a while at least just until she got something else sorted out.\\
& \textit{Hypothesis:} she could stay \\
& Heuristics: \texttt{D} \hspace{0.2cm} V. BERT: \texttt{D} \hspace{0.2cm} J. BERT: \texttt{D} \hspace{0.2cm} AAs: \texttt{N} \{\texttt{N}, \texttt{N}, \texttt{N}\}  \\ 
& Gold: \texttt{N} [3, 0, 0, 0, 0, 0, 0, 0, 0, 0]  \\ \midrule
3 & \textit{Premise:} A: but that is one of my solutions. Uh... B: I know here in Dallas that they have just instituted in the last couple of years, uh, a real long period of time that you can absentee vote before the elections. And I do not think they have seen a really high improvement.\\
& \textit{Hypothesis:} they have seen a really high improvement. \\
& Heuristics: \texttt{C} \hspace{0.2cm} V. BERT: \texttt{C} \hspace{0.2cm} J. BERT: \texttt{C} \hspace{0.2cm}  AAs: \texttt{C} \{\texttt{C}, \texttt{C}, \texttt{C}\}  \\ 
 & Gold: \texttt{C} [-1, -2, -2, -2, -2, -2, -2, -2, -3, -3] \\ \midrule
4 & \textit{Premise:}B: So did you commute everyday then or, A: No. B: Oh, okay. A: No, no, it was a six hour drive. B: Oh, okay, when you said it was quite a way away, I did not know that meant you had to drive like an hour \\
& \textit{Hypothesis:} speaker A had to drive like an hour\\
& Heuristics: \texttt{C} \hspace{0.2cm} V. BERT: \texttt{D} \hspace{0.2cm} J. BERT: \texttt{E} \hspace{0.2cm}  AAs: \texttt{D} \{\texttt{E}, \texttt{C}, \texttt{C}\}  \\ 
 & Gold: \texttt{D} [3, 2, 2, 1, 0, 0, -1, -1, -1, -3]\\  \midrule
 5 & \textit{Premise:} The assassin's tone and bearing were completely confident. If he noticed that Zukov was now edging further to the side widening the arc of fire he did not appear to be troubled. \\
& \textit{Hypothesis:} Zukov was edging further to the side\\
& Heuristics: \texttt{D} \hspace{0.2cm} V. BERT: \texttt{D} \hspace{0.2cm} J. BERT: \texttt{D} \hspace{0.2cm}  AAs: \texttt{D} \{\texttt{E}, \texttt{E}, \texttt{N}\}  \\ 
& Gold: \texttt{E} [3, 3, 3, 3, 2, 2, 1, 1]\\ \midrule
 6 & \textit{Premise:} B: Yeah, and EDS is very particular about this, hair cuts, A: Wow. B: I mean it was like you can't have, you know, such and such facial hair, no beards, you know, and just really detailed. A: A: I don't know that that would be a good environment to work in. \\
& \textit{Hypothesis:} that would be a good environment to work in \\
& Heuristics: \texttt{C} \hspace{0.2cm} V. BERT: \texttt{C} \hspace{0.2cm} J. BERT: \texttt{D} \hspace{0.2cm} AAs: \texttt{C} \{\texttt{C}, \texttt{C}, \texttt{C}\}  \\ 
& Gold: \texttt{D} [2, 0, 0, 0, 0, -1, -2, -3] \\ \midrule
7 & \textit{Premise:} ``Willy did mention it. I was puzzled, I 'll admit, but now I understand.'' How did you know Heather had been there? \\
& \textit{Hypothesis:} Heather had been there \\
& Heuristics: \texttt{N} \hspace{0.2cm} V. BERT: \texttt{E} \hspace{0.2cm} J. BERT: \texttt{E} \hspace{0.2cm} AAs: \texttt{E} \{\texttt{E}, \texttt{E}, \texttt{E}\}  \\ 
& Gold: \texttt{D} [3, 3, 3, 2, 1, 1, 0, 0, 0] \\ 
 \bottomrule
\end{tabular}
}
\caption[Models' predictions for CB test items.]{Models' predictions for CB test items. Labels in \texttt{[]} are predictions by individual AAs. }
\label{tbl_ch4:qualitative}
\end{table}

%% file: Table/Table_ch4_8.tex
\begin{table}
\centering
\resizebox{0.5\linewidth}{!}{%
\begin{tabular}{l|rrrr|r}
\toprule
\multicolumn{1}{c}{\backslashbox{Predict}{Gold}}  & \texttt{E} & \texttt{N} & \texttt{C} & \texttt{D} & Total  \\ \cmidrule(lr){2-6}
\texttt{E} & 37 & 2 & 0 & 13 & 52\\
\texttt{N} &  1 & 10 & 0 & 3 & 14\\
\texttt{C} & 0 & 0 & 34 & 13 & 47\\
\texttt{D} & 20 & 7 & 20 & 80 & 127 \\
\midrule
Total &  58 &  19 & 54 & 109 & 240 \\ \bottomrule
\end{tabular}
}
\caption[Confusion matrix for the test set.]{Confusion matrix for the test set. \texttt{E}: entailment, \texttt{N}: neutral, \texttt{C}: contradiction, \texttt{D}: disagreement. }
\label{tbl_ch4:error_analysis}
\end{table}

%% file: Table/Table_ch4_5.tex
\begin{table}
\centering
\resizebox{0.6\linewidth}{!}{%
\begin{tabular}{lrrrrr}
\toprule
 & negation & modal & conditional & question & negR \\\midrule
Heuristic & 51.29 & 48.02 & 37.69 & 44.64 & 54.16 \\
V. BERT & 60.91 & 73.98 & 44.84 & 53.02 & 61.91 \\
J. BERT & 60.94 & 73.95 & 46.02 & 51.68 & 63.67 \\\midrule
AAs & \textbf{65.96} & \textbf{80.18} & \textbf{48.05} & \textbf{54.95} & \textbf{68.00} \\ \bottomrule
\end{tabular}
}
\caption{F1 for CB test set under the embedding environments and ``I don't know/believe/think'' (``negR").}
\label{tbl_ch4:additional}
\end{table}

%% file: Table/Table_ch4_6.tex
\begin{table}
\centering
\resizebox{0.6\linewidth}{!}{%
\begin{tabular}{lrrrr}
\toprule
\multicolumn{1}{c}{\multirow{2}{*}{\pbox{20cm}{Correct inference\\ by Heuristic?}}}  & \multicolumn{2}{c}{Yes (130)} & \multicolumn{2}{c}{No (110)} \\ 
\cmidrule(lr){2-3}
\cmidrule(lr){4-5}
\multicolumn{1}{c}{} & Acc. & F1 & Acc. & F1 \\ \midrule
V. BERT & 80.00 & 80.45 & 41.51 & 42.48 \\
J. BERT & 79.74 & 80.04 & 42.73 & 44.15 \\
AAs & \textbf{84.37} & \textbf{84.85} & \textbf{46.97} & \textbf{48.75}\\ \bottomrule
\end{tabular}
}
\caption[BERT-based models performance on test items correctly predicted by vs.\ items missed by linguistic rules.]{BERT-based models performance on test items correctly predicted by vs.\ items missed by linguistic rules. Numbers next to Yes/No denote the size.}
\label{tbl_ch4:linguistics}
\end{table}

%% file: ch3.FAQ.tex
\chapter{FAQ Retrieval
\label{FAQ}}

\section{Introduction}
FAQ, short for frequently asked questions, is designed for the purpose of providing information on frequent questions or concerns. The FAQ retrieval task is defined as ranking FAQ items $\{(q_i,a_i)\}$ from an FAQ Bank given a user query $Q$. In the FAQ retrieval literature \citep{Karan2016, Karan2018, FAQ-Sakata}, a user query $Q$ can be learned to match with the question field $q_i$, the answer field $a_i$ or their concatenation (i.e., FAQ tuple) $q_i+a_i$.

\begin{figure}[t]
    \centering
    \includegraphics[width=1.0\linewidth]{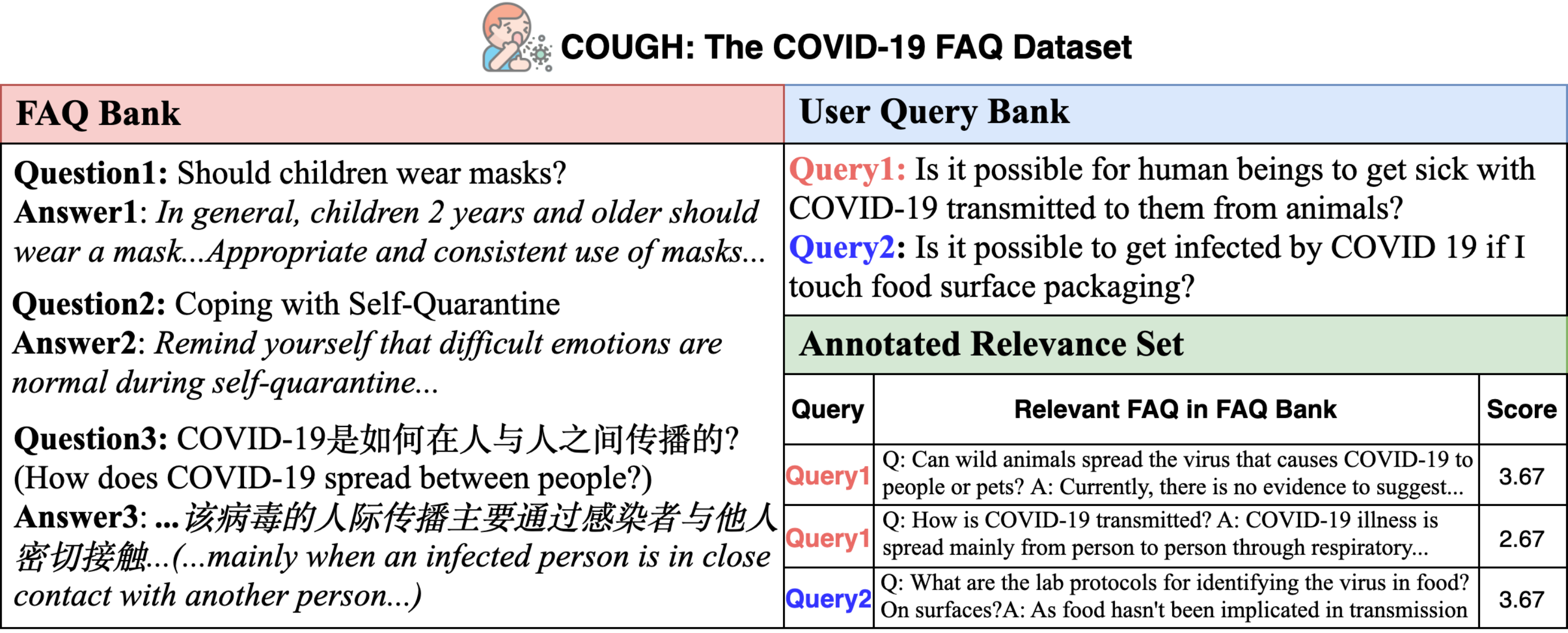}
    \caption{Examples from the \cough dataset.}
    \label{fig_ch3:intro_eg}
\end{figure}

To advance the COVID-19 information search, we present an FAQ dataset, \cough, consisting  of FAQ Bank, Query Bank, and Relevance Set, following the standard of constructing a robust FAQ dataset \citep{Manning-IR}. The FAQ  Bank  contains 15919  FAQ  items scraped from 55 authoritative institutional websites. \cough covers a wide range of perspectives on COVID-19, spanning from general information about the virus to specific COVID-19-related instructions for a healthy diet. For evaluation, we further construct Query  Bank  and    Relevance  Set, including 1201 crowd-sourced queries and their relevance to a set of FAQ items judged by annotators. Examples from \cough are shown in Figure \ref{fig_ch3:intro_eg}.

Our dataset poses several new challenges (e.g., the answers being long and noisy, and hard to match due to larger search space) to existing FAQ retrieval models. The diversity of FAQ items, which is reflected in their varying query forms and lengths as well as in narrative styles, also contributes to these challenges. Furthermore, these challenges can reflect the characteristics and difficulties of FAQ retrieval in real scenarios better than counterparts like FAQIR \citep{Karan2016} and StackFAQ \citep{Karan2018} (Table~\ref{tbl_ch3:compare}). Moreover, in contrast to all prior datasets, \cough covers multiple query forms (e.g., question and query string forms) and has many annotated FAQs for each user query, whereas queries in existing FAQ datasets are limited to the question form and have much fewer annotations. As such, our \cough is deemed as a robust dataset, upon which a robust FAQ retriever could be developed to handle some real challenges (e.g., lengthy answer, enormous search space) better. 

The contribution in this chapter is two-fold. First, we construct a challenging dataset \cough to aid the development of COVID-19 FAQ retrieval models. Second, we conduct extensive experiments using various SOTA models across different settings, explore limitations of current FAQ retrieval models, and discuss future work along this line.

\section{Standard FAQ Dataset Construction: \cough}

Since the outbreak of COVID-19, the community has witnessed many datasets released to advance the research of COVID-19.The  most  related  work  to  ours are \citet{covid-Sun} and \citet{covid-Poliak}, both of which constructed a collection of COVID-19 FAQs by scraping authoritative websites (e.g., CDC and WHO). However, the dataset in the former work is not available yet and the latter work does not evaluate models on their dataset, and there is still a great need to understand how existing models would perform on the COVID-19 FAQ retrieval task. 
Moreover, the numbers of FAQs\footnote{In the literature, only 789 FAQ items are used for evaluation on FAQIR \citep{Karan2018, Unsupervised-FAQ}.} in the 5 existing FAQ datasets (Table~\ref{tbl_ch3:compare}) are generally lower than 2000, which renders a small search space and thus the ease for FAQ retrievers to find the most relevant FAQ given a query.  
\input{Table/Table_ch3_2}

A typical research-oriented FAQ dataset \citep{Manning-IR} consists of three parts: FAQ Bank, User Query Bank and Annotated Relevance Set. In this section, we will describe how we construct each of the three in detail.

\subsection{FAQ Bank Construction}
We developed scrapers based on JHU-COVID-QA library \citet{covid-Poliak} with modifications to enable special features for our \cough dataset.

\noindent\textbf{Web scraping:} We collect FAQ items from authoritative international organizations, state governments and some other credible websites including reliable encyclopedias and medical forums. 
Moreover, we scrape three types of FAQs: question form (i.e., an interrogative statement), query string (i.e., a string of words to elicit information) form and forum form (FAQs scrapped from medical forums). 
Inspired by \citet{Manning-IR}, we loosen the constraint that queries must be in question form since we want to study a more generic and challenging problem. We also scrape 6,768 non-English FAQs to increase language diversity. Overall, we scraped a total of 15,919 FAQ items covering all three types and 19 languages. All FAQ items were collected and finalized on Aug. 30$^{th}$, 2020. 

\subsection{User Query Bank Construction}
\label{userQueryBankConstruction}

Following \citet{Karan2016,Manning-IR}, we do not crowdsource queries from scratch, but instead ask annotators to paraphrase our provided query templates (See phase 1 below for details). In this way, we can ensure that 1) the collected queries are pertinent to COVID-19; 2) the collected queries are not too simple; 3) the chance of getting (nearly) duplicate user queries is reduced. 

\noindent\textbf{Phase 1: Query Template Creation}: We sample 5\% of FAQ items from each English non-forum source and use the question part as the template.

\noindent\textbf{Phase 2: Paraphrasing for Queries}: In this phase, each annotator is expected to give three paraphrases for each query template. Annotators are encouraged to give deep paraphrases (i.e.,  grammatically different but semantically similar/same) to simulate the noisy and diverse environment in real scenarios. In the end, we obtain 1236 user queries.

\subsection{Annotated Relevance Set Construction}
\label{AnnotatedRelevanceSetConstruction}
\noindent\textbf{Phase 1: Initial Candidate Pool Construction}: For each user query, as suggested by previous work \citep{Manning-IR,Karan2016, FAQ-Sakata}, we run 4 models\footnote{Explanations of these models are in chapter \ref{Unsupervised_FAQ_Retrieval}.}, BM25 (Q-q), BM25 (Q-q+a), BERT (Q-q) and BERT (Q-a) fine-tuned on \cough, to instantiate a candidate FAQ pool.  Each model complements the others and contributes its top-10 relevant FAQ items. We then take the union to remove duplicates, giving an average pool size of 32.2.

\noindent\textbf{Phase 2: Human Annotation}: Each annotator gives each $\langle$Query, FAQ item$\rangle$ tuple a score based on the annotation scheme (i.e., Matched (4), Useful (3), Useless (2) and Non-relevant (1)) which is adapted from \citet{Karan2016, FAQ-Sakata}. In order to reduce the variance and bias in annotation, each tuple has at least 3 annotation scores.
In our finalized Annotated Relevance Set, we keep all raw scores and include two additional labels: 1) mean of raw annotation scores; 2) binary label (positive/negative). We identify all tuples with mean score greater than 3 as positive examples.

Among 1236 user queries, we find that there are 35 ``unanswerable" queries that have no associated positive FAQ item. In the end, there are 1201 user queries involved for evaluation after removing ``unanswerable" queries.

\section{\cough Dataset Analysis}
\label{data-analysis}

Besides the generic goal of large size, diversity, and low noise, COUGH features 4 additional aspects:

\input{Table/Table_ch3_3}

\noindent \textbf{Varying Query Forms:}
As indicated in Table \ref{tbl_ch3:data}, there are multiple query forms. In evaluation, we include both question and query string forms. These two distinct forms are  different in terms of query format (interrogative vs. declarative), average answer length (123.89 vs. 89.60) and topics. Question form is usually related to general information about the virus while query string form is often searching for more specific instructions concerning COVID-19 (e.g., healthy diet during pandemic). In Figure \ref{fig_ch3:intro_eg}, the first FAQ item is in question form while the second one is in query string form.

\noindent \textbf{Answer Nature:}
Table \ref{tbl_ch3:compare} shows the answer fields in \cough are much longer than those in any prior dataset. We also observe that answers might contain some contents which are not directly pertinent to the query, partially resulting in the long length nature. For example, in \cough, the answer to a query ``What is novel coronavirus" contains extra information about comparisons with other viruses. Such lengthy and noisy nature of answers shows the difficulty of FAQ retrieval in real scenarios.

\noindent \textbf{Large-scale Relevance Annotation:} 
Many existing FAQ datasets overlooked the scale of annotations (Table \ref{tbl_ch3:compare}); yet, that would hurt the  evaluation reliability since many true positive $\langle$Query, FAQ item$\rangle$ tuples were omitted. Following \citet{Manning-IR}, for each user query, we constructed a large-scale candidate pool to reduce the chance of missing true positive tuples. The annotation procedure yielded 39760 annotated $\langle$Query, FAQ item$\rangle$  tuples, each of which is annotated by at least 3 people to reduce annotation bias. Furthermore, we find that there are 7856 (19.76\%) positive tuples (i.e., mean score $>$ 3). Besides, from the perspective of FAQ Bank, 6648 of 7117 English non-forum items appear at least once in Initial Candidate Pool, and 3790 of them have at least one ``matched" user query.

\noindent \textbf{Multilinguality:} 
\cough includes 6768 FAQ items covering 18 non-English languages. In this thesis, we do not include FAQ items in languages other than English in the evaluation.\footnote{No annotation is done on non-Engligh items.} However, we do encourage investigators who use \cough to better utilize non-English FAQ items for other potential tasks, such as multi-lingual FAQ retrieval and transfer learning from English FAQ items to low-resource non-English FAQ items.

\begin{figure}[t]
  \centering
  \includegraphics[width=0.8\textwidth]{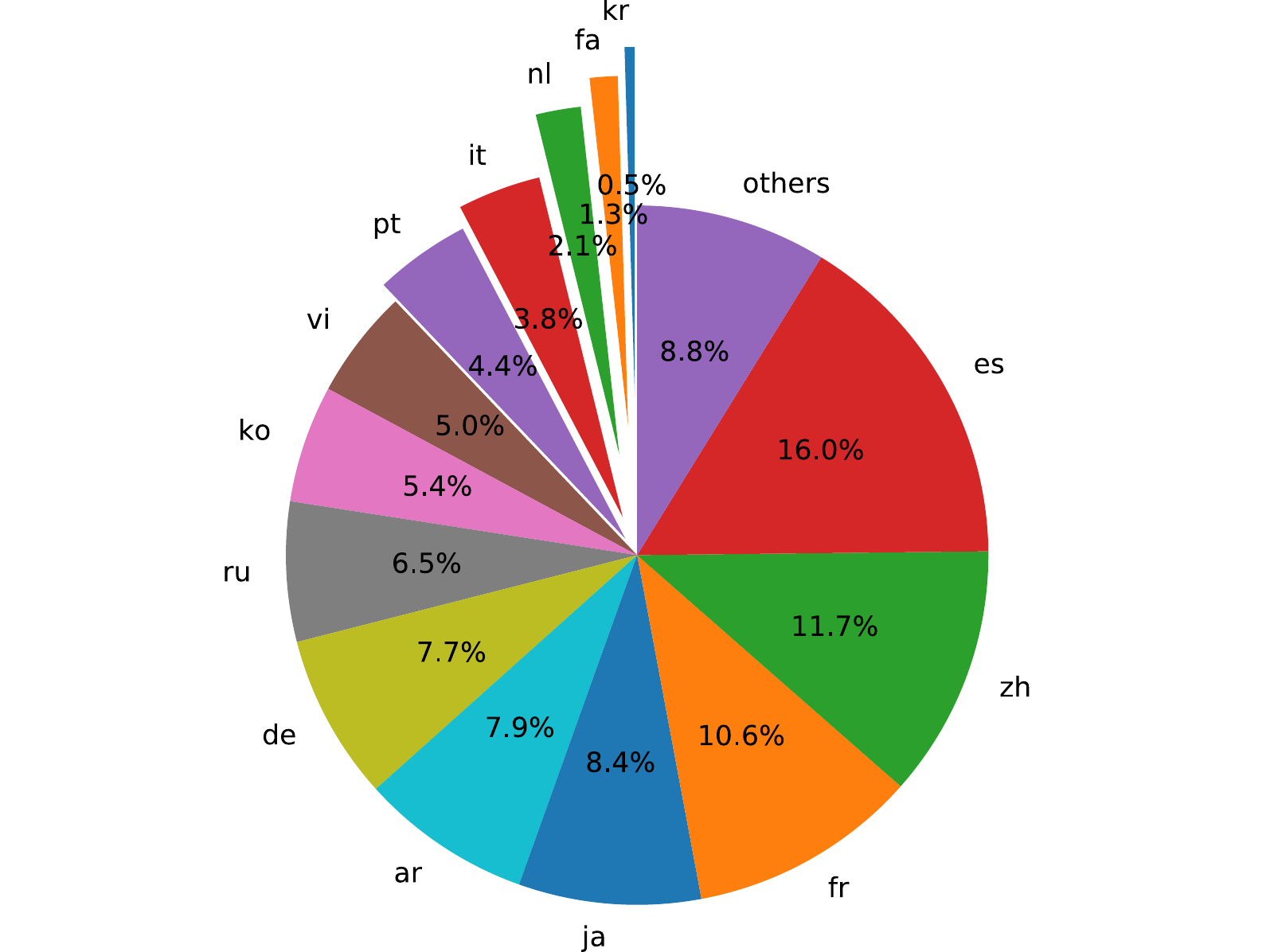}
  \caption{Language distribution for non-English FAQ items.}
  \label{fig_ch3:language}
\end{figure}

Figure \ref{fig_ch3:language} shows the language distribution (excluding English) of FAQ items in \cough dataset. Like English FAQ items, non-English FAQ items are also presented in both question and query string forms. 
Statistics of non-English items can be found in Table~\ref{tbl_ch3:data}.

\section{FAQ Retrieval Methods}

\subsection{FAQ Retrieval Methods Overview}

The standard practice in FAQ retrieval focuses on retrieving the most-matched FAQ items given a user query \citep{Karan2018}. 
Many earlier work, such as FAQ FINDER \citep{Burke-1997}, query expansion \citep{KIM2006650} and BM25 \citep{INR-019}, resorted to traditional IR techniques by leveraging lexical mapping and/or semantic similarity. In the deep learning era, many studies show that Neural Networks are useful for FAQ retrieval as they are good at learning the semantic relevance between queries and FAQ items. 
Along this line, \citep{Karan2016} adopted Convolution Neural Networks, \citep{Gupta-LSTM} utilized LSTM, and \citep{FAQ-Sakata} leveraged an ensemble of TSUBAKI \citep{TSUBAKI} and BERT \citep{devlin2019bert}. 
Recently, \citet{Unsupervised-FAQ} employed CombSum and PoolRank, ensembles of BM25 and BERT models,
to learn ranking without requiring manual annotations.

\subsection{Unsupervised FAQ Retrieval}
\label{Unsupervised_FAQ_Retrieval}

In this chapter, we only focus on the unsupervised models since the size of User Query Bank (1201 items) is not large enough for supervised learning, especially for fine-tuning complex language models like BERT. We experiment with three commonly-used and SOTA unsupervised models to understand their limitations and figure out the challenge present in real scenarios for FAQ retrieval. Besides, each model has three configurable modes, Q-q, Q-a and Q-q+a, where we match user queries (Q) to the question (q) and answer (a) of an FAQ item as well as their concatenation (q+a)\footnote{Q-q+a mode is only used for BM25 and BM25 in CombSum.}, respectively.

\subsubsection{Baseline Models}

\noindent (1) \textbf{BM25} \citep{INR-019}, a commonly adopted IR baseline, is a nonlinear combination of term frequency, document frequency and document length.

\vspace{1mm}
\noindent (2) \textbf{BERT} \citep{devlin2019bert} is a pretrained language model. We experiment with Sentence-BERT \citep{sentence-bert}, a Siamese network built for comparison between sentence-pair embeddings,  which specializes in generating meaningful sentence representations.

\noindent \textbf{Fine-tuning:} We  use Multiple Negatives Ranking (MNR) loss\footnote{For efficiency, MNR loss is computed using answers of other FAQs in the same training batch as negative answers.} \citep{henderson2017efficient} to fine-tune Sentence-BERT on FAQ bank.
For the Q-q mode, similar to \citet{Unsupervised-FAQ}, we use GPT2 \citep{radford2019language} to generate synthetic questions as positive q's to match with Q and filter out low-quality ones via Elasticsearch.
For the Q-a mode, an FAQ item itself is a positive pair. For both modes, negative q's or a's are randomly sampled. 

\vspace{1mm}

\noindent (3) \textbf{CombSum} \citep{Unsupervised-FAQ}
first computes three matching scores between the user query and FAQ items via BM25 (Q-q+a), BERT (Q-q) and BERT (Q-a) models, respectively. Then, the three scores are normalized and combined by averaging.

\section{Evaluation and Results}

\noindent \textbf{Evaluation Metric:} We adopt our binary label (positive/negative) as ground truth labels. Following previous work \citep{Karan2016,Karan2018,FAQ-Sakata}, we adopt  widely-used MAP (Mean Average Precision)\footnote{Evaluated on top-100 retrieved FAQ items.}, MRR (Mean Reciprocal Rank) and P@5 (Precision at top 5) metrics.

\noindent \textbf{Evaluation Settings:}  For the scope of this chapter, we only evaluate on English non-forum FAQ items, and leave the non-English and forum ones for future research as great challenges have already been observed under the current setting. However, we do encourage investigators who use \cough to utilize these two categories for other potential applications (e.g., multi-lingual IR, transfer learning in IR).

\input{Table/Table_ch3_4}

\noindent \textbf{Evaluation Results:} Models' results are listed in Table~\ref{tbl_ch3:result}. The current best results (P@5: 0.31; MAP: 0.42; MRR: 0.64) are not satisfying, showing a large room for improvement. These results not only confirm that \cough is challenging but also signify more robust methods and models are needed to handle challenges imposed by \cough more effectively.

\section{Analysis}

\noindent \textbf{Quantitative Analysis:}
It is not surprising to see that the Q-q mode consistently performs better than the Q-a mode regardless of underlying models. This is mainly caused by the fact that question fields are more similar to user queries than answer fields, in terms of syntactic structures and semantic meanings. As discussed in Section \ref{data-analysis}, the answer nature (lengthy and noisy) and large search space, albeit well characterize the FAQ retrieval task in real scenarios, do bring a great challenge to current FAQ retrieval models.

We observe that fine-tuning in the way we experimented with can only help improve the performance of the Q-a mode by a small margin, but might slightly hurt the Q-q mode due to the noise introduced in generating synthetic queries. Moreover, ensemble models don't perform as well as expected, since the particular Q-a model involved is weak (even after fine-tuning), which negatively impacts performance. In consequence, doing straightforward fine-tuning or ensemble simply by stacking models wouldn't improve the performance significantly, which confirms that \cough is a challenging dataset. Interesting future work includes developing more advanced techniques to handle long and noisy answer fields.

\input{Table/Table_ch3_5}

\noindent \textbf{Qualitative Analysis:} To understand finetuned BERT (Q-q) better, we conduct error analysis as shown in Table~\ref{tbl_ch3:error} to show its major types of errors, hoping to further improve it in the future. Currently, finetuned BERT (Q-q) suffers from the following issues: 1) biased towards responses with similar texts (e.g., “antibody tests” and “antibody testing”); 2) fails to capture the semantic similarities under complex environments (e.g., pragmatic reasoning is required to understand that “limited ability” indicates results are not accurate for diagnosing COVID-19).

\section{Conclusion}

In this chapter, we introduce \cough, a large challenging dataset for COVID-19 FAQ retrieval. \cough features varying query forms, long and noisy answers, larger search space and multilinguality. \cough also serves as a better evaluation benchmark since it has large-scale relevance annotations. Albeit results show the limitations of current FAQ retrieval models, \cough is a more robust dataset than its counterparts since it better characterizes the challenges present in real scenarios for FAQ retrieval.

%% file: Table/Table_ch3_2.tex
\begin{table*}[t]
\centering
\resizebox{\linewidth}{!}{%
\begin{tabular}{lrrrrrr}
\toprule
 & \pbox{20cm}{\textbf{FAQIR}\\(\citeauthor{Karan2016})}  & \pbox{20cm}{\textbf{StackFAQ}\\(\citeauthor{Karan2018})} & \pbox{20cm}{\textbf{LocalGov}\\(\citeauthor{FAQ-Sakata})} & \textbf{\citeauthor{covid-Sun}}  & \textbf{\citeauthor{covid-Poliak}}  & \textbf{\cough (ours)} \\ \midrule
Domain & Yahoo! & StackExchange & Government & COVID-19 & COVID-19 & COVID-19 \\
\# of FAQs & 4,313 & 719 & 1,786 & 690 & 2,115 & 15,919 \\
\# of Queries (Q) & 1,233 & 1,249 & 784 & 6,495* & 24,240* & 1,201 \\
\# of annotations per Q & 8.22 & Not Applicable & $<$10 & 5 & 5 & 32.17 \\ \midrule
Query Length & 7.30 & 13.84 & ** & ** & ** & 12.97 \\
FAQ-query Length & 12.30 & 10.39 & ** & ** & ** & 13.00 \\
FAQ-answer Length & 33.00 & 76.54 & ** & ** & ** & 113.58 \\ \midrule
Language & English & English & Japanese & English & Multi-lingual & Multi-lingual \\
\# of sources & 1 & 1 & 1 & 12 & 34 & 55 \\ \bottomrule
\end{tabular}
}
\caption[Comparison of \cough with representative counterparts.]{Comparison of \cough with representative counterparts. *: Extracted from existing resources (e.g., COVID-19 Twitter dataset \citep{Chen_2020}).  **: Not Applicable, either not in English or not publicly available.}
\label{tbl_ch3:compare}
\end{table*}

%% file: Table/Table_ch3_3.tex
\begin{table}[t]
\centering
\resizebox{0.6\linewidth}{!}{%
\begin{tabular}{lcccc}
\toprule
 & Type &Number & Q-Length & A-length\\ 
 \midrule
 \multirow{3}*{\# English}
    & Question & 4978 & 14.64 & 123.89  \\ 
    & Query String  & 2139 & 9.18 & 89.60  \\ 
    & Forum & 2034 & 147.46 & 90.49  \\ 
\midrule

  \multirow{2}*{\# Non-English}
     &  Question &3396  & - & - \\ 
     & Query String  &3372 & - & -  \\
     
\midrule
\# Total & - &15919  & - & - \\ \bottomrule
\end{tabular}
}
\caption{Basic statistics of FAQ bank in \cough.}
\label{tbl_ch3:data}
\end{table}

%% file: Table/Table_ch3_4.tex
\begin{table}[t]
\centering
\resizebox{0.6\linewidth}{!}{
\begin{tabular}{lccc}
\toprule
 & P@5 & MAP & MRR \\ \midrule
BM25 (Q-q) & 0.27 & 0.38 & 0.56 \\
BM25 (Q-a) & 0.16 & 0.23 & 0.34 \\ 
BM25 (Q-q+a) & 0.25 & 0.34 & 0.52 \\ \midrule
BERT (Q-q) w/o finetune & 0.29 & \textbf{0.42} & 0.59 \\
\, + finetune on pesudo Q-q & 0.26 & 0.36 & 0.60 \\ 
BERT (Q-a) w/o finetune & 0.06 & 0.12 & 0.17 \\
\, + finetune on FAQ Bank & 0.23 & 0.30 & 0.50 \\ \midrule
CombSum w/o finetune & 0.21  & 0.31 & 0.49 \\
\, + fintune on pesudo Q-q & 0.23 &0.31 &0.53 \\
\, + fintune on FAQ Bank & 0.31 &0.39 &0.63 \\
\, + fintune on pesudo Q-q and FAQ Bank & \textbf{0.31} & 0.39 & \textbf{0.64} \\\bottomrule
\end{tabular}
}
\caption[Evaluation on \cough.]{Evaluation on \cough. BERT refers to Sentence-BERT \citep{sentence-bert}.}
\label{tbl_ch3:result}
\end{table}

%% file: Table/Table_ch3_5.tex
\begin{table}[t]
\resizebox{\linewidth}{!}{%
\begin{tabular}{p{1.2\linewidth}l}
\toprule
\textbf{Query:} What research is being done on antibody tests and their accuracy?  \\
\textbf{FAQ item:} Q: What is antibody testing? How do I get a COVID-19 antibody test? A: CDC and partners are investigating to determine if you can get sick with COVID-19 more than once ...\\
\textbf{Gold label:} \texttt{Negative} \texttt{[}useful, useless, useless\texttt{]}\\ 
\textbf{Predicted rank:} \texttt{3}\\ \midrule
\textbf{Query:} Are COVID-19 antibody tests accurate?  \\
\textbf{FAQ item:}  Q: Should I be tested with an antibody (serology) test for COVID-19? A: ... Antibody tests have limited ability to diagnose COVID-19 and should not be used alone to diagnose COVID-19 ... \\
\textbf{Gold label:} \texttt{Positive} \texttt{[}useful, useful, matched\texttt{]} \\ \textbf{Predicted rank:} \texttt{26} \\ \bottomrule
\end{tabular}
}
\caption[Error analysis with fine-tuned BERT (Q-q).]{Error analysis with fine-tuned BERT (Q-q).  Human annotations are inside \texttt{[]}.}
\label{tbl_ch3:error}
\end{table}

%% file: ch5.Conclusion.tex
\chapter{Conclusion}
\label{conclusion}

In this thesis, I have embarked on building models and constructing datasets towards more robust natural language understanding. We start with a discussion on what robustness problem is in natural language understanding. That is, fully-trained NLU models are usually lacking generalizability and flexibility. In this thesis, we argue that, in order to achieve truly robust natural language understanding, implementing robust models and curating robust datasets are equally important. In this thesis, we investigate the NLU robustness problem in three NLU tasks (i.e., Question Answering, Natural Language Inference and Information Retrieval). We then propose novel methods and construct new datasets to advance research on improving the robustness of NLU systems. 

In Chapter~\ref{CQA}, we study how to utilize diversity boosters (e.g., beam search \& QPP) to help Question Generator synthesize diverse QA pairs, upon which a Question Answering (QA) system is trained to improve the generalization onto unseen target domain. It's worth mentioning that our proposed QPP (question phrase prediction) module, which predicts a set of valid question phrases given an answer evidence, plays an important role in improving the cross-domain generalizability for QA systems. Besides, a target-domain test set is constructed and approved by the community to help evaluate the model robustness under cross-domain generalization setting. In Chapter~\ref{NLI}, we investigate inherently ambiguous items in the NLI (Natural Language Inference) task, which are overlooked in the literature but often occurring in the real world, for which annotators don’t agree with the gold label. We build an ensemble model, AAs (Artificial Annotators), which simulates underlying annotation distribution to effectively identify such inherently ambiguous items. Our AAs, motivated by the nature of inherently ambiguous items, are better than vanilla models since our model design captures the essence of the problem better.  In Chapter~\ref{FAQ}, we follow a standard practice to build a robust dataset for FAQ retrieval task. In our dataset analysis, we show how \cough better reflects the challenge of FAQ retrieval in the real situation than its counterparts. The imposed challenge (e.g., long and noisy answer, large search space) will push forward the boundary of research on FAQ retrieval in real scenarios.

Overall, the technical contributions of this thesis are as follows:

\begin{itemize}[leftmargin=2.8em,noitemsep,topsep=0pt,parsep=0pt,partopsep=0pt]
    \item[1. ] We investigate the robustness problem in depth, and identify the equal importance of models implementation and datasets construction towards improving the robustness of NLU systems. In this thesis, we specifically study three concrete NLU tasks.
    \item[2. ] We propose two novel methods to help improve NLU model robustness. Specifically, we evaluate the effect of diverse question generation (QG) for clinical QA under the cross-domain evaluation setting, and propose QPP (Question Phrase Prediction) module as an effective diversity booster for QG \citep{cliniqg4qa2020}. Moreover, we propose AAs (Artificial Annotators) to simulate underlying annotation distribution to handle a previously-overlooked NLI class better, inherent disagreement items \citep{zhang2021NLI}.
    \item[3. ] We construct two robust datasets, QA test set on MIMIC-III Database \citep{cliniqg4qa2020} and COUGH \citep{zhang2020cough}. They will serve as better evaluation benchmarks to examine designed models' generalization capabilities and abilities to handle real-scenario challenges (e.g., longer FAQ and larger search space).
\end{itemize}

\noindent\textbf{Future Research:}
Moving forward, the ultimate goal for robust natural language understanding is to build NLU models which can behave humanly. That is, it's expected that robust NLU models are capable to transfer the knowledge from training corpus to unseen documents more reliably and survive when encountering challenging items even if the model doesn't know a priori of users' inputs. Two suggested important research frontiers are:

\noindent\textbf{1)  Improve model generalization under cross-domain setting:} In Chapter~\ref{CQA}, we discussed how we utilized QG model to help alleviate the generalization challenge encountered by QA systems. However, the question whether a better QA system could further improve the QG is yet known, which is, however, worth deeper investigation. Ideally, when introducing an auxiliary module to help the main model, we also expect to see that the auxiliary module could be benefited by the joint training with the main model. Besides, in Chapter~\ref{CQA}, the reason we decided to utilize QG that way is that we observed that the QG system didn't suffer from severe generalization issues under the clinical setting. However, in open-domain, the aforementioned observation might not hold. In that case, it might be better to enforce the model to learn text representations that are invariant to domain changes. Recent work on cross-domain NER (Named Entity Recognition) have shown some progress along this path \citep{JiaXZ19}.
I also have a great interest in text generation. Though the majority of work that utilize domain adaptation techniques to tackle the generalization challenge focuses on classification tasks \citep{GaninUAGLLML16,ChenC18,ChenSACW18}, could we effectively extend the success of domain adaptation to text generation? This might be a promising research direction since the text generation can be formulated as a sequence of classifications.

\noindent\textbf{2) Embrace more challenges in NLU:}
In Chapter~\ref{NLI} and \ref{FAQ}, we discussed two datasets, CommitmentBank \& \cough, on which we could develop methods that target at solving NLU challenges under more realistic scenarios. SQuAD 2.0 \citep{rajpurkar2018know} is a another great role model for datasets that aim at this goal. To do well on SQuAD 2.0, models must not only answer questions when possible, but also determine when no answer is supported by the paragraph and then say ``no''. This is a real challenge for QA system as it's not always the case that an answer could be found in a seemingly relevant document for a question. Another typical real challenge in NLU is how to solve mathematical problems. \citet{NLP_Math} presents a new math dataset on which a standard CS PhD student who doesn't especially like Math gets 40\% accuracy while a fully-trained GPT-3 \citep{BrownMRSKDNSSAA20} models only gets ~5\%. Pretrained language models like GPT-3 or BERT is believed to heavily rely on the context to reason about the given prompt. However, mathematical language isn't necessarily constrained by contexts,\footnote{Math question could be context-free such as "let \textit{a} equal one plus two minus three times four, is \textit{a} congruent to zero when the modulo is five?"} which imposes a great challenge to NLU systems. Additionally, in order to get full credits for a problem, the deployed system is also required to give correct reasoning steps, which is way more difficult than simply generating an answer. The following is an example from MATH dataset:\footnote{This example corresponds to the second example in their Figure 1.}
\begin{itemize}[leftmargin=6em,noitemsep,topsep=0pt,parsep=0pt,partopsep=0pt]
    \item[\textit{Problem:} ] \textit{If $\Sigma_{n=0}^{\infty}cos^{2n}\theta=5$, what is $cos2\theta$?}
    \item[\textit{Solution:} ] \textit{The geometric series is $1+cos^2\theta+cos^4\theta+...=\frac{1}{1-cos^2\theta}=5$. Hence, $cos^2\theta=\frac{4}{5}$. Then, $cos2\theta=2cos^{2}\theta-1=\frac{3}{5}$}
\end{itemize}
Moreover, linguistic rules or features, without any doubt, deserve more attention even if we are living in the realm of neural computing world.  This is because linguistic rules or features exhibit great power when tackling challenging NLU problems. In Chapter~\ref{NLI}, we find that SOTA NLU models, BERT, obtain inferior results to our linguistics-driven heuristic rules on dev set. This shows that giant neural models still fail to capture some necessary linguistic phenomena. As such, it's essential to discover how to effectively incorporate linguistic information into neural models to compensate for what the neural network-model is weak at. A simple practice is to embed linguistic features such as NER and POS tags into original texts. Particularly, I observed that a vanilla attention-based Seq2Seq model, when being equipped with linguistic features, could achieve better performance than BART \citep{LewisBART},\footnote{In general, BART ($\sim$139M) has 8 times more parameters than vanilla attention-based Seq2Seq ($\sim$17M).} a variant of BERT specializing in text generation, on both  in-domain and cross-domain question generation tasks.

%% file: app1.tex
\chapter{Supplementary Materials}

\section{Clinical Question Answering}
\subsection{Answer Evidence Extractor}
\label{apx:extract_evidence}

\noindent\textbf{Formulation and Implementation}
Formally, given a document (context) $\vec{p}=\{p_1,p_2,...,p_m\}$, where $p_i$ is the $i$-th token of the document and $m$ is the total number of tokens, we aim to extract potential evidence sequences. Firstly, we adopt the ClinicalBERT model \citep{alsentzer2019publicly}
to encode the document:
\vspace{-6pt}
\begin{equation}
\vspace{-2pt}
    \mathbf{U} = \texttt{ClinicalBERT} \{p_1,...,p_m\}.
\end{equation}
where $\mathbf{U} \in \mathbb{R}^{m\times d}$, and $d$ is size of the dimension.

Following the same paradigm of the BERT model for the sequence labeling task \citep{devlin2019bert}, we predict the \texttt{BIO} tag for each $a_j$ as follows:
\vspace{-5pt}
\begin{equation}
\vspace{-3pt}
    \Pr (a_j|p_i)=\text{softmax}(\mathbf{U} \cdot \mathbf{W} + \mathbf{b}), \;\forall p_i\in \vec{p}
\end{equation}

We train model on source contexts by minimizing the negative log-likelihood loss.

\noindent\textbf{Post-processing Heuristic Rules}
We observe that when we directly apply the ClinicalBERT \citep{alsentzer2019publicly} system described in Section~\ref{Overview_of_Our_Framework} on clinical texts, the extracted answer evidences sometimes are broken sentences due to the noisy nature and uninformative language (e.g., acronyms) of clinical texts. To make sure the extracted evidences are meaningful, we designed a \textit{``merge-and-drop''} heuristic rule to further improve the extractor’s accuracy. Specifically, for each extracted evidence candidate, we first examine the \textit{length}  (number of tokens) of the extracted evidence. If the length is larger than the threshold $\eta$, we keep this evidence; otherwise, we compute the \textit{distance}, i.e., the number of tokens between the current candidate span and the closest span. If the \textit{distance} is smaller than the threshold $\gamma$, we merge these two ``close-sitting'' spans; otherwise, we drop this over-short evidence span. In our experiments, we set $\eta$ and $\gamma$ to be 3 and 3, respectively, since they help the QA system achieve the best performance on the dev set.

\subsection{Question Phrases Identification}
\label{apx:QPP_I}
In order to utilize the Question Phrase Prediction (QPP) module and make the QPP module generic enough without loss of generality, we identify valid n-gram Question Phrases in an automatic way.

To prepare an exhaustive list of valid n-gram Question Phrases, we first collect all of the first \textit{n} words appearing in questions in emrQA, forming three (i.e., \textit{n}=1, 2, 3) raw Question Phrases set.

We observe that all uni-grams are valid question phrases (e.g., ``How'', ``When'', ``What''), so we don't do any pruning and keep the uni-gram question phrases set as it is.

As for n-gram (n $\geq$ 2) Question Phrases set, we conduct fine-grained filtering. We only consider n-grams with occurrence frequency greater than the threshold $\zeta$ as valid n-gram Question Phrases. In our experiment, we set $\zeta$ as 0.02\%. Less frequent n-gram words (i.e., frequency $<$ 0.02\%) will degrade to unigram Question Phrases in accordance with corresponding question types (e.g., ``Has lasix'' $\rightarrow$ ``Has''*) so as to maintain lossless. In the end, n-gram (n $\geq$ 2) Question Phrases sets, without any information loss, consist of both n-gram Question Phrases and degraded unigram Question Phrases.

\subsection{Dev Set Construction}
\label{DevSetConstruction}
The dev set on MIMIC-III is constructed by sampling generated questions from 9 QG models and is used to tune the hyper-parameters only. Instead of uniformly sampling from 9 QG models, we followed the sampling ratio of 1:3:6 (Base model, Base+BeamSearch, Base+QPP) for each QG method, which made the dev set cover as many diverse questions as possible.